\documentclass[sigconf]{acmart} 

\usepackage{subfigure}
\usepackage{graphicx}
\usepackage{bbm}
\usepackage{multirow}
\usepackage{bbding}
\usepackage{amsmath}

\usepackage{algorithm}
\usepackage{algorithmic}
\usepackage{amsmath}

\usepackage{color}

\newcommand{\tabincell}[2]{\begin{tabular}{@{}#1@{}}#2\end{tabular}}

\usepackage{colortbl}
\definecolor{light-gray}{gray}{0.90}

\AtBeginDocument{%
  }

\copyrightyear{2023}
\acmYear{2023}
\setcopyright{acmlicensed}
\acmConference[KDD '23] {Proceedings of the 29th ACM SIGKDD Conference on Knowledge Discovery and Data Mining}{August 6--10, 2023}{Long Beach, CA, USA.}
\acmBooktitle{Proceedings of the 29th ACM SIGKDD Conference on Knowledge Discovery and Data Mining (KDD '23), August 6--10, 2023, Long Beach, CA, USA}
\acmPrice{15.00}
\acmISBN{979-8-4007-0103-0/23/08}
\acmDOI{10.1145/3580305.3599374}

\begin{document}
\title{GraphSHA: Synthesizing Harder Samples for Class-Imbalanced Node Classification}

\author{Wen-Zhi Li}
\affiliation{%
  \institution{CSE, Sun Yat-sen University, Guangzhou, China}
  \city{}\country{}
}
\affiliation{%
  \institution{AI Thrust, HKUST (GZ), Guangzhou, China}
  \city{}\country{}
}
\email{liwzh63@mail2.sysu.edu.cn}

\author{Chang-Dong Wang}
\authornote{Corresponding authors.}
\affiliation{%
  \institution{CSE, Sun Yat-sen University}
  \city{Guangzhou}\country{China}
}
\email{changdongwang@hotmail.com}

\author{Hui Xiong}
\authornotemark[1]
\affiliation{%
  \institution{AI Thrust, HKUST (GZ), Guangzhou, China}
  \city{}\country{}
}
\affiliation{%
  \institution{CSE, HKUST, Hong Kong, China}
  \city{}\country{}
}
\email{xionghui@ust.hk}

\author{Jian-Huang Lai}
\affiliation{%
  \institution{CSE, Sun Yat-sen University}
  \city{Guangzhou}\country{China}
}
\email{stsljh@mail.sysu.edu.cn}

\begin{abstract}
Class imbalance is the phenomenon that some classes have much fewer instances than others, which is ubiquitous in real-world graph-structured scenarios. Recent studies find that off-the-shelf Graph Neural Networks (GNNs) would under-represent minor class samples.
We investigate this phenomenon and discover that the subspaces of minor classes being squeezed by those of the major ones in the latent space is the main cause of this failure. We are naturally inspired to enlarge the decision boundaries of minor classes and propose a general framework GraphSHA by \underline{S}ynthesizing \underline{HA}rder minor samples. Furthermore, to avoid the enlarged minor boundary violating the subspaces of neighbor classes, we also propose a module called \textsc{SemiMixup} to transmit enlarged boundary information to the interior of the minor classes while blocking information propagation from minor classes to neighbor classes. 
Empirically, GraphSHA shows its effectiveness in enlarging the decision boundaries of minor classes, as it outperforms various baseline methods in class-imbalanced node classification with different GNN backbone encoders over seven public benchmark datasets.
Code is avilable at \url{https://github.com/wenzhilics/GraphSHA}.
\end{abstract}

\ccsdesc[500]{Computing methodologies~Learning latent representations}
\ccsdesc[500]{Mathematics of computing~Graph algorithms}

\keywords{node classification; class imbalance; graph neural network; hard sample; data augmentation}


\maketitle

\begin{figure}[!t]
\centering
\includegraphics[width=0.95\linewidth]{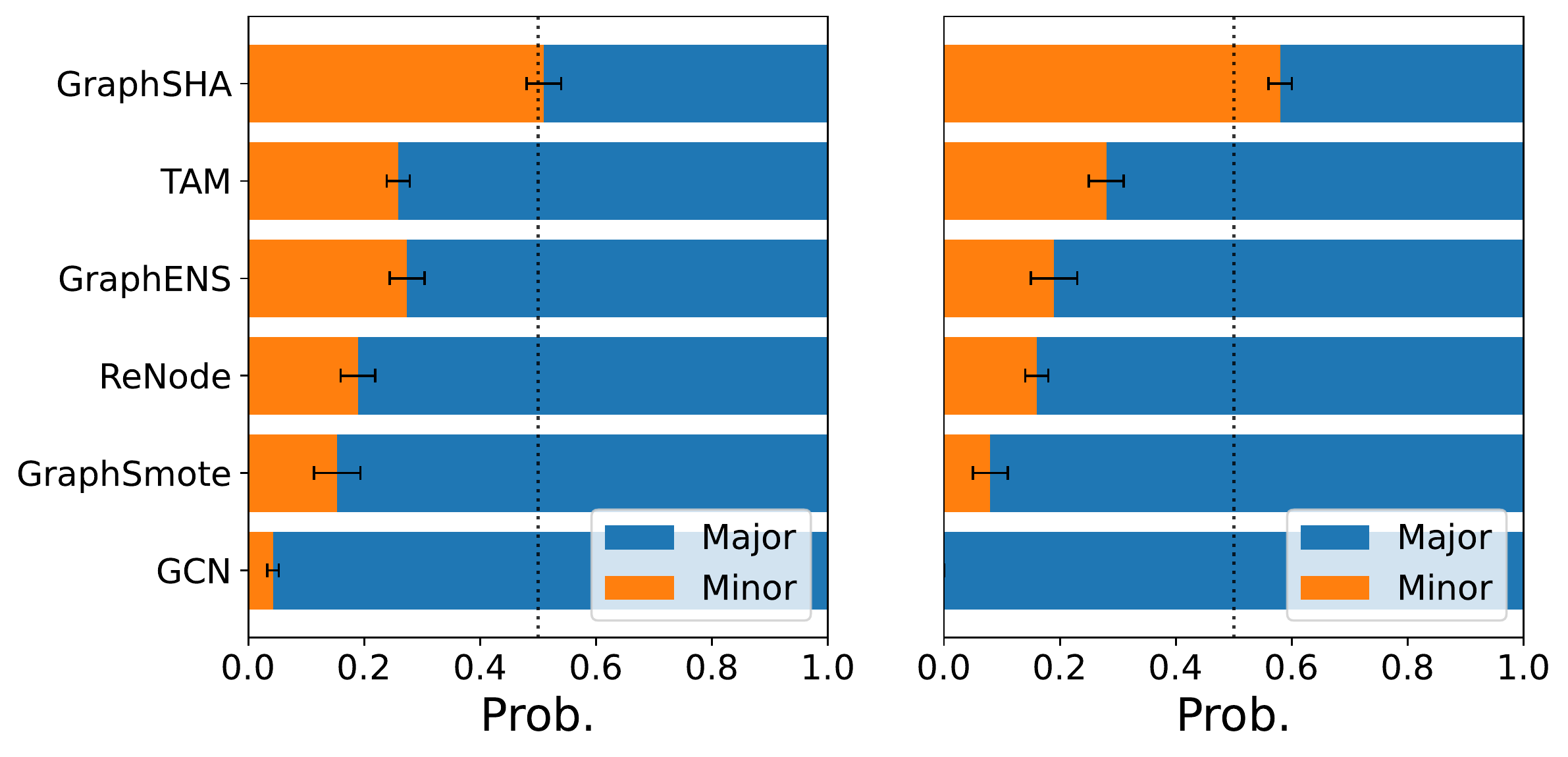}
\centering
\vspace{-0.1cm}
\caption{Probability distribution of misclassified samples on Cora-LT (left) and CiteSeer-LT (right) datasets w.r.t. different methods. Both datasets are preprocessed to follow a long-tailed distribution with an imbalance ratio of 100 as in~\cite{graphens}. We treat the half classes with fewer instances as minor ones and the other half as major ones. We can see that GCN suffers from the \textit{squeezed minority problem}, as nearly all false samples are classified as major classes. Though baseline methods GraphSmote, ReNode, GraphENS, and TAM can remit this problem to some extent, the minor subspaces are still squeezed as their distributions are still highly biased. Our GraphSHA, on the other hand, can significantly enalrge the minor subspaces as the probability of misclassified samples being minor classes is close to 0.5 (black dotted line).}\label{figure:smp}
\vspace{-0.7cm}
\end{figure}

\vspace{-0.5cm}
\section{Introduction}

Node classification is regarded as a vital task for graph analysis~\cite{graphmae, cook2006mining}. With the fast development of neural networks in the past few years, Graph Neural Networks (GNNs) have become the de-facto standard to handle this task and achieved remarkable performances in many graph-related scenarios~\cite{gcn, gat, gnn1, gnn2}. Current GNNs are implicitly based on the class balance assumption where the numbers of training instances in all classes are roughly balanced~\cite{graphsmote}. This assumption, however, barely holds for graph data in-the-wild as they tend to be \textit{class-imbalanced} intrinsically~\cite{graphens, renode}.
In class-imbalanced graphs, some classes (minor ones) possess much fewer instances than others (major ones). For example, in a large-scale citation network~\cite{mag}, there are more papers on artificial intelligence than on cryptography. 
Applying GNNs to these class-imbalanced graphs would under-represent minor classes, leading to non-ideal overall classification performance~\cite{graphsmote, tam}.

\vspace{-0.2cm}
\subsubsection*{\textbf{Reseach Background}} 
Like imbalance handling methods in other domains, there are generally two aspects of work adapting GNNs to class-imbalanced graphs.
Generative approaches~\cite{graphsmote, drgcn, imgagn, gial} aim to augment the original class-imbalanced graph by synthesizing plausible minor nodes to make the class distribution balanced,
while loss-modifying approaches~\cite{renode, tailgnn, tam} aim to adjust the objective function to pay more attention to minor class samples.
We focus on generative approaches in this paper as topology information can also be synthesized to leverage the intrinsic property of graph structure, thus leading to superior performance empirically~\cite{graphsmote, graphens}.

We revisit the class-imbalanced issue on graphs from the latent space perspective with an empirical study in Figure~\ref{figure:smp}, which shows that GNNs suffer from the \textit{squeezed minority problem}, where the latent space determined by imbalanced training data is highly skewed — minor subspaces are squeezed by major ones.
As a result, the minor test samples are hard to be included into their correct latent subspaces, leading to unsatisfactory classification performance during the inference phase eventually.

Faced with the squeezed minority problem, we argue that the goal of generative approaches is to \textit{enlarge the minor decision boundaries in the latent space}. With this insight, we are naturally inspired to leverage hard samples that determine the decision boundary for the synthesis. That is, if we can synthesize \textit{harder} minor samples beyond the hard minor ones, the squeezed minority problem can thus be alleviated.

\vspace{-0.2cm}
\subsubsection*{\textbf{Main Challenge}} 
Clear though the goal is, enlarging minor class boundary is not trivial. As the boundary is shared by a minor class and its neighbor class,
naively synthesizing harder minor samples would unavoidably violate the neighbor class subspace, thus degenerating neighbor class, which is apparently not what we expect. As a countermeasure, a proper augmentation method is required to \textit{enlarge the subspaces of minor classes while avoiding deteriorating those of the neighbor ones}.

To tackle this challenge, we propose a fine-grained augmentation approach, GraphSHA, for \underline{S}ynthesizing \underline{HA}rder minor samples.
Specifically, the synthesis is based on an anchor node and an auxiliary node, where the anchor node is a hard minor sample, and the auxiliary node is sampled from the anchor node's neighbor class.
To remit degenerating neighbor classes, we propose a novel module called \textsc{SemiMixup} for the synthesis of the harder minor sample where its feature is generated via the mixup~\cite{mixup} between the features of the anchor node and the auxiliary node, while the edges connecting it are only sampled from the anchor node's 1-hop subgraph without mixup.
As the auxiliary node's 1-hop subgraph whose nodes belong to neighbor class with high confidence according to graph homophily~\cite{homo1, homo2} is excluded, \textsc{SemiMixup} can disable information propagation from the minor class to the neighbor class while enabling the propagation of the information beyond the raw minor boundary to the interior of the minor class, which can enlarge the decision boundary of minor class properly without deteriorating neighbor class.

We validate our method on various real-world benchmark datasets, including citation networks Cora, CiteSeer, PubMed~\cite{cora}, co-purchase networks Amazon-Photo, Amazon-Computers~\cite{amazon}, and co-author network Coauthor-CS~\cite{amazon} in both long-tailed~\cite{graphens} and step class imbalance settings~\cite{renode} with diverse GNN backbones including GCN~\cite{gcn}, GAT~\cite{gat}, and GraphSAGE~\cite{graphsage}. We also conduct experiments on a large-scale naturally class-imbalanced dataset ogbn-arXiv~\cite{ogb}. 
The experimental results show the generalizability of GraphSHA and the effectiveness of the \textsc{SemiMixup} module in enlarging the decision boundaries of minor classes by avoiding deteriorating the subspaces of neighbor ones.

To sum up, we highlight the main contributions of this work as:

\begin{itemize}
\item We find that the squeezed minority problem, where the subspaces of minor classes are squeezed by major ones, is the main cause for the unsatisfactory performance of minor classes in class-imbalanced node classification.
\item To enlarge the squeezed minor subspaces, we propose a novel generative method, GraphSHA, to \underline{S}ynthesize \underline{HA}rder minor samples. Moreover, we propose a module called \textsc{SemiMixup} as the key component of GraphSHA to avert invading neighbor subspaces when enlarging the minor subspaces.
\item Extensive experiments and in-depth analysis demonstrate the effectiveness of GraphSHA and the \textsc{SemiMixup} module, as GraphSHA consistently outperforms state-of-the-art baseline methods across various public benchmark datasets.
\end{itemize}

\section{Related Work}\label{sec:related}

In this section, we briefly review the prior work related to this paper, including class imbalance handling methods and hard sample mining methods.

\subsection{Class Imbalance Problem}

The class imbalance problem is widespread in real-world applications for various machine learning tasks~\cite{classimbalance1, classimbalance2}. As major classes have much more samples than minor classes in the training set, machine learning models are believed to easily under-represent minor classes, which results in poor overall classification results~\cite{tam}.

Existing countermeasures to remit the class imbalance problem can be roughly divided into two categories: loss-modifying and generative approaches.
Loss-modifying approaches~\cite{reweight11, reweight22, reweight33} are devoted to modifying the loss function to focus more on minor classes. Generative approaches~\cite{generative11, generative22, generative33} are devoted to generating minor samples to balance the training set.
Directly applying these approaches to graph data cannot achieve satisfactory results as graphs possess edges between node samples intrinsically~\cite{graphsmote, tam}.

\begin{figure}[!t]
\centering
\includegraphics[width=0.99\linewidth]{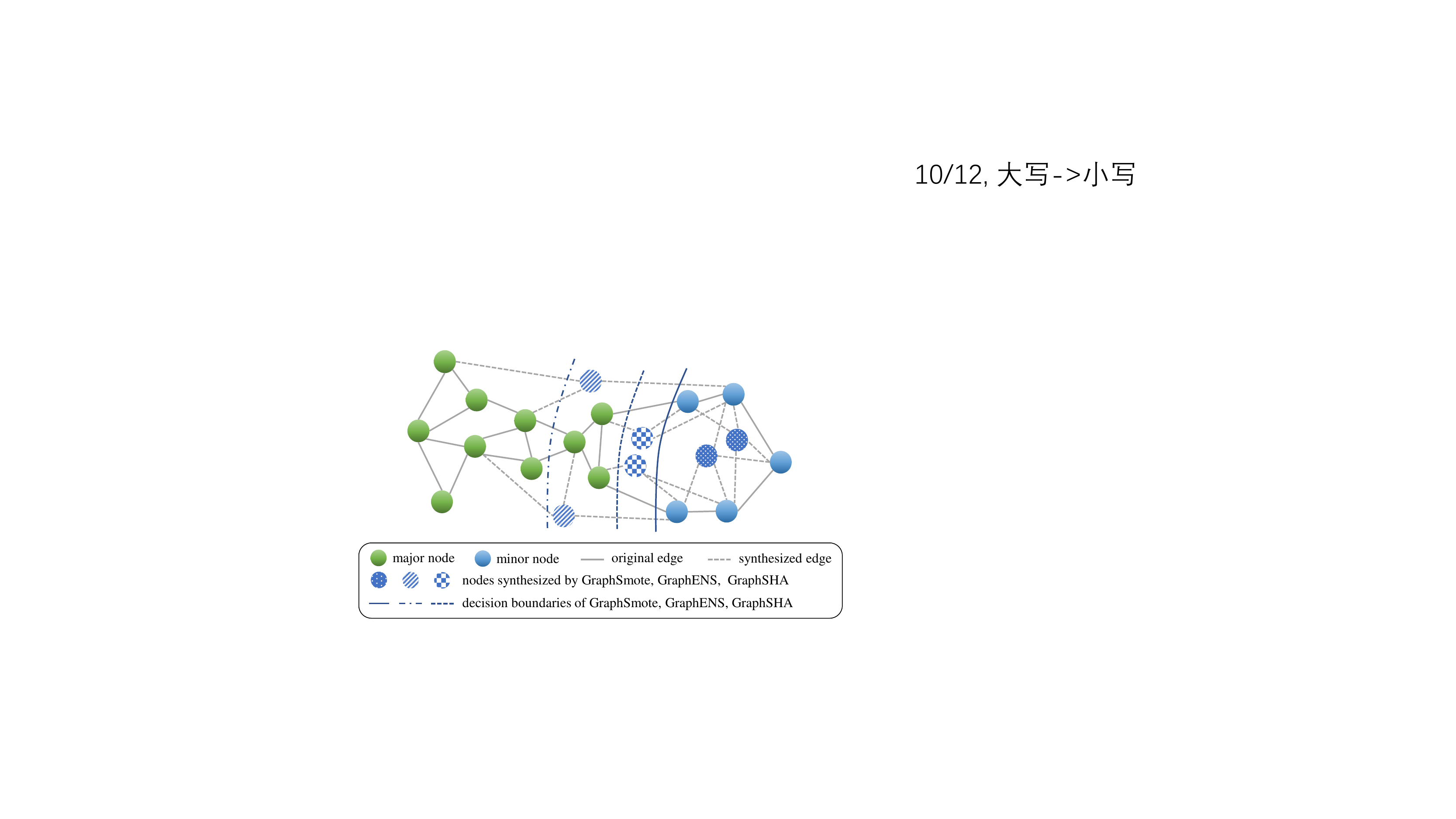}
\centering
\vspace{-0.1cm}
\caption{Comparison of the synthesis for GraphSmote~\cite{graphsmote}, GraphENS~\cite{graphens}, and GraphSHA. GraphSmote can only generate minor nodes within the subspace, which is not beneficial to alleviating the squeezed minority problem. GraphENS, on the other hand, generates minor nodes far beyond the decision boundary, which degenerates major class accuracy. Different from these methods, GraphSHA generates minor nodes beyond the minor decision boundary properly, which can effectively enlarge the subspace of minor class.}\label{figure:compare}
\vspace{-0.2cm}
\end{figure}

For graph data, methods to handle class-imbalanced node classification are primarily generative approaches, as edges can be synthesized together with node samples. Among them, GraphSmote~\cite{graphsmote} synthesizes minor nodes by interpolating between two minor nodes in the same class in the SMOTE~\cite{smote} manner, and an extra edge predictor is leveraged to generate edges for the synthesized nodes. DR-GCN~\cite{drgcn} and ImGAGN~\cite{imgagn} leverage neural network GAN~\cite{gan} to synthesize minor nodes. However, ImGAGN is only capable of binary classification tasks of distinguishing minor nodes from major ones, which is non-trivial to be applied to multi-label classification tasks in this paper. For these methods, the synthesized nodes are generated based on existing minor nodes, which are still confined to the raw minor subspaces and bring limited gain to the squeezed minor class. GraphENS~\cite{graphens}, on the other hand, synthesizes ego networks for minor classes by combining an ego network centered on a minor sample and another one centered on a random sample from the entire graph, which certainly enlarges the decision boundary of minor classes. However, the heuristic generation overdoes the minor node generation, thus unavoidably increasing false positives for major classes. To sum up, these generative methods cannot enlarge the minor subspaces effectively and precisely to alleviate the bias, which is the main focus of this work. The comparison of GraphSmote, GraphENS, and GraphSHA is illustrated in Figure~\ref{figure:compare}.

It is worth mentioning that there are also loss-modifying approaches for graph data like ReNode~\cite{renode} and TAM~\cite{tam}. However, they mainly focus on the topological imbalance issue, i.e., the imbalanced connectivity of nodes in the graph, which is beyond the scope of our work. Nevertheless, we also compare these methods in Section~\ref{sec:exp}.

\subsection{Hard Sample Mining}

Hard samples, i.e., samples that are difficult for the current model to discriminate, are believed to play a crucial role in classification tasks~\cite{focal}.
They are often leveraged in self-supervised learning, where the objectives are roughly defined as ``maximizing the similarity between positive pairs while minimizing the similarity between negative pairs''~\cite{simclr}. As positive pairs are often limited while negative pairs are exhaustive, hard sample mining is typically referred to as hard negative mining, to name a few~\cite{mochi, hcl, dcl, ring, progcl, mmix}. Similar to the \textsc{SemiMixup} module in our work, FaceNet~\cite{facenet} also discovers that naively selecting hard negative samples in face recognition would practically lead to local minima, thus introducing semi-hard samples. However, it differs from our work as we consider data synthesis (FaceNet only chooses among existing negative samples) in graph data (FaceNet deals with image data).

In the scope of class-imbalanced classification, some pioneering work also leverage hard samples~\cite{focal, imbhard1, imbhard2} to avoid easy negatives overwhelming the training phase. However, they are generally loss-modifying methods. As mentioned above, they cannot exploit the topology information, which makes them hard to be applied to class-imbalanced graphs. On the other hand, GraphSHA synthesizes hard samples in the graph domain with edges, which enables it to tackle the problem naturally.

\section{Preliminaries}

\subsection{Notations and Imbalance Settings}

We focus on semi-supervised node classification task on an unweighted and undirected graph $\mathcal{G=\{V,E \}}$, where $\mathcal{V}=\{v_1,\cdots,v_N\}$ is the node set with $N$ nodes and $\mathcal{E} \subseteq \mathcal{V} \times \mathcal{V}$ is the edge set. The adjacency matrix and the feature matrix are denoted as $\boldsymbol{A} \in \{ 0,1 \}^{N \times N}$ and $\boldsymbol{X} \in \mathbb{R}^{N \times d}$ respectively, where $\boldsymbol{A}_{ij}=1$ iff $(v_i,v_j)\in\mathcal{E}$, and $\boldsymbol{X}_i\in\mathbb{R}^d$ is a $d$-dim raw feature of node $v_i$. $\mathcal{N}_i$ is the direct neighbor set of node $v_i$.

Every node $v$ corresponds with a class label $\boldsymbol{Y}(v) \in \{1,\cdots,C\}$ with $C$ classes in total, and we denote all nodes in class $c$ as $\boldsymbol{Y}_c$.
In class-imbalanced node classification, the labeled nodes in training set $\mathcal{V}^L\subset\mathcal{V}$ are imbalanced, where the imbalance ratio is defined as $\rho = \max_i |\boldsymbol{Y}_{i}| / \min_j |\boldsymbol{Y}_{j}|$. 
 
\subsection{Graph Neural Networks}

GNNs are the yardsticks for feature extraction of graph-based literature. They generally follow the ``propagation-transformation'' paradigm to iteratively fuse neighbor node features as:
\begin{equation}
\boldsymbol{H}_{t}^{(l)} \leftarrow \boldsymbol{\operatorname { Transform }}\left(\underset{\forall v_s \in \mathcal{N}_t}{\boldsymbol{\operatorname { Propagate }}}\left(\boldsymbol{H}_{s}^{(l-1)} ; \boldsymbol{H}_{t}^{(l-1)}\right)\right),
\end{equation}
where $\boldsymbol{H}_{t}^{(l)}$ is the node embedding of node $v_t$ in the $l$-th layer.
For example, a two-layer GCN~\cite{gcn} can be formalized as
\begin{equation}
\begin{aligned}
&\boldsymbol{\operatorname { Propagate }}: \boldsymbol{M}_t^{(l)}=\boldsymbol{\hat{A}}\boldsymbol{H}_t^{(l-1)},\\
&\boldsymbol{\operatorname { Transform }}: \boldsymbol{H}_t^{(l)}=ReLU(\boldsymbol{M}_t^{(l)}\boldsymbol{W}^{(l)}),
\end{aligned}
\end{equation}
where $\boldsymbol{\hat{A}}$ is the normalized adjacency matrix, $\boldsymbol{W}^{(l)}$ is learnable weight matrix of the $t$-th layer, and $ReLU(\cdot)$ is the ReLU activation function. We also consider several other GNN variants in this paper, which vary from the $\boldsymbol{\operatorname { Propagate }}(\cdot)$ and $\boldsymbol{\operatorname { Transform }}(\cdot)$ functions including GAT~\cite{gat} and GraphSAGE~\cite{graphsage}.

\subsection{Problem Definition}

The goal of generative class imbalance handling approaches is to augment the raw imbalanced graph $\mathcal{G}$ by synthesizing minor nodes (including node features and edges) to make it balanced. Then, the augmented balanced graph $\mathcal{G}^{\prime}$ is fed into a GNN encoder $f_{\theta}(\cdot)$ afterwards for traditional GNN-based node classification task. 

\begin{figure*}[!t]
\centering
\includegraphics[width=1.0\linewidth]{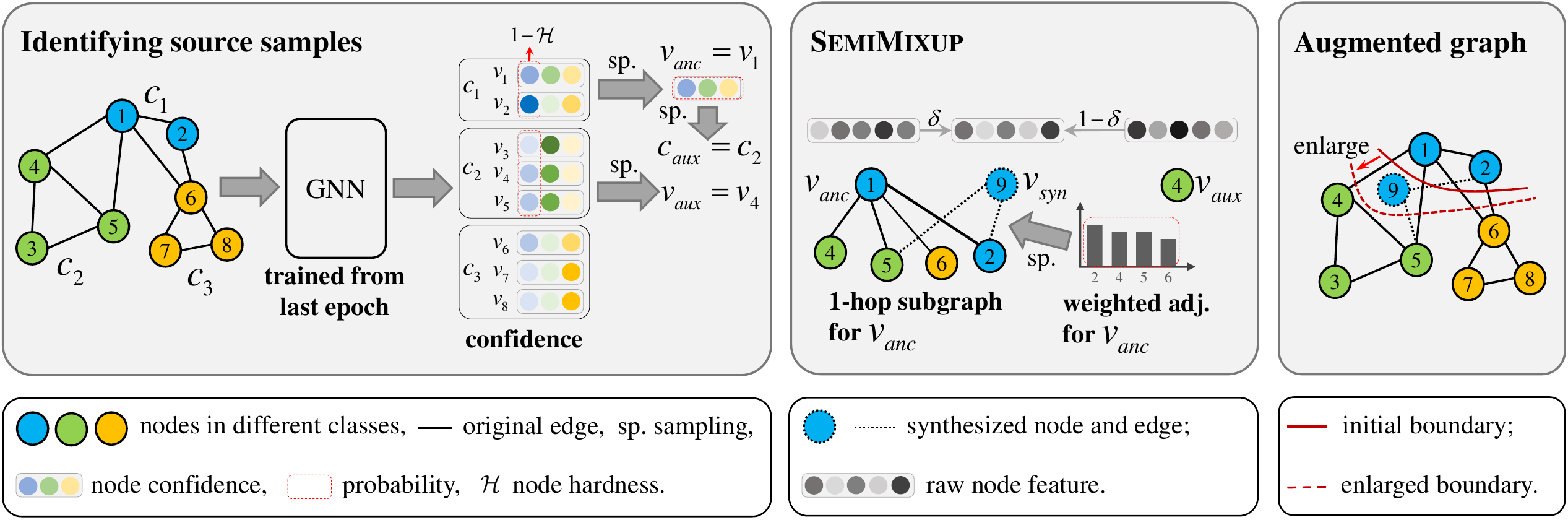}
\centering
\caption{
GraphSHA overview where $c_1$ is minor class and $c_2$, $c_3$ are major classes. 
\textbf{(Left)}: two source nodes $v_{anc}$ and $v_{aux}$ are firstly identified via three samplings: sampling from minor nodes in $c_1$ according to their hardness $\mathcal{H}$ to get $v_{anc}$; sampling from major classes $c_2$, $c_3$ according to $v_{anc}$'s confidence on them to get neighbor class $c_{aux}$; and sampling from nodes in neighbor class $c_{aux}$ according to their confidences on minor class $c_1$ to get $v_{aux}$. 
\textbf{(Middle)}: $\textsc{SemiMixup}$ is conducted by mixuping $v_{anc}$'s 1-hop subgraph and $v_{aux}$ solely to get synthesized node $v_{syn}$. 
Here, the unweighted subgraph is transformed into a weighted one via diffusion-based smoothing based on graph topology. 
\textbf{(Right)}: the augmented graph is fed into a GNN for traditional node classification task where the minor class decision boundary is enlarged properly without degenerating neighbor classes.
}
\label{figure:flowchart}
\end{figure*}

\section{Methodologies}
In this section, we introduce the proposed GraphSHA model in detail.
We first describe each component of GraphSHA. Then, we provide the complexity analysis. The overall framework is illustrated in Figure~\ref{figure:flowchart}, and the training algorithm is elaborated in Algorithm~\ref{algo:graphsha}.

\subsection{Identifying Source Samples}\label{sec:identifying}

We are motivated to enlarge the minor decision boundary, which is determined by hard minor anchor node $v_{anc}$. We also need an auxiliary node $v_{aux}$ from $v_{anc}$'s neighbors in the latent space so that the boundary is enlarged from $v_{anc}$ towards $v_{aux}$.

Many methods can be leveraged to calculate node hardness in the latent space, such as confidence~\cite{graphconf} and $K$-Nearest Neighbor ($K$NN). Without loss of generalizability, we adopt confidence as the hardness metric for its light-weighted computational overhead. Extra experiments with $K$NN-based node hardness are in Appendix~\ref{appendix:exp}.

\begin{definition}[confidence-based node hardness]
For a $C$-shot classification task, let $\boldsymbol{Z}_i \in \mathbb{R}^{C}$ be the logits for node $v_i$, i.e., $\boldsymbol{Z}_i=f_{\theta}(v_i)$. The hardness for node $v_i$ is defined as
\begin{equation}\label{eq:hardness}
\mathcal{H}_i=1-\sigma_{SM}\left(\boldsymbol{Z}_{i,\boldsymbol{Y}(v_i)}\right),
\end{equation}
where $\sigma_{SM}(\boldsymbol{Z}_{i,\cdot})=\frac{\exp (\boldsymbol{Z}_{i,\cdot}/T)}{\sum_{j=1}^{C}\exp(\boldsymbol{Z}_{i,j}/T)}$ is the softmax function with temperature $T$~\cite{distill}.
\end{definition}

In practice, node logits $\boldsymbol{Z}^{\prime}$ from the previous epoch can be leveraged to get their hardness $\mathcal{H}$. Thus we can identify minor anchor nodes $v_{anc}$ by sampling from a multinomial distribution with the hardness as the probability.

In identifying auxiliary node $v_{aux}$, we first determine the neighbor class $c_{aux}$ of each anchor node by sampling from a multinomial distribution with $\sigma_{SM}(\boldsymbol{Z}_{anc}$) as the probability. 
Then, for nodes in $c_{aux}$, we sample from another multinomial distribution with their confidence in class $\boldsymbol{Y}(v_{anc})$ as the probability. In this way, we can get two souce nodes $v_{anc}$ and $v_{aux}$ near the boundary of minor class.
An example in identifying $v_{anc}$ and $v_{aux}$ is given on the left side of Figure~\ref{figure:flowchart}.

\subsection{\textsc{SemiMixup} for Harder Sample Synthesis}
Based on the two source nodes $v_{anc}$ and $v_{aux}$, harder minor node $v_{syn}$ involves two folds of synthesis: node feature synthesis and edge synthesis. The goal of the synthesized node is to enlarge the minor class boundary while avoiding degenerating neighbor class in the latent space, which is implemented via \textsc{SemiMixup}.

\subsubsection*{\textbf{Synthesizing Node Features}} 

The raw feature of $v_{syn}$ can be generated via a simple mixup~\citep{mixup} between node embeddings of $v_{anc}$ and $v_{aux}$ in the raw feature space as 
\begin{equation}\label{eq:mix}
\boldsymbol{X}_{syn}=\delta\boldsymbol{X}_{anc}+(1-\delta)\boldsymbol{X}_{aux},\; \delta\in [0,1].
\end{equation}
Here, smaller $\delta$ will force the generated node feature to be more analogous to the auxiliary node, which is expected to be more beneficial to enlarge the minor decision boundary. We validate this via an empirical study in Section~\ref{sec:hyper} as smaller $\mathbb{E}(\delta)$ will contribute to better model performance.

\subsubsection*{\textbf{Synthesizing Edges}} 

As features are propagated via edges in GNN, we expect the synthesized edges to enable propagating information beyond the minor class boundary to the interior of the minor class while blocking propagation from the minor class to the neighbor class to avoid degenerating the neighbor class. 
To this end, instead of connecting $v_{syn}$ with the nodes in the \textit{union} of $v_{anc}$'s 1-hop subgraph and $v_{aux}$'s 1-hop subgraph, we make a simple adaptation by only connecting $v_{syn}$ with the nodes in $v_{anc}$'s 1-hop subgraph, as only $v_{anc}$'s 1-hop subgraph tend to share the same minor label with $v_{anc}$ according to graph homophily~\cite{homo1, homo2}. In this way, message passing in GNNs would enable the enlarged boundary information — represented as the synthesized node feature — to be propagated to the interior of the minor class.

However, $v_{anc}$'s 1-hop sugraph may contain less nodes than we want to sample. Furthermore,
as the graph is unweighted, uniform sampling from the subgraph may ignore important topology information.
Our countermeasure is to transform the unweighted hard graph into a weighted soft one based solely on graph topology.
Here, we refer to graph diffusion-based smoothing proposed in GDC~\cite{gdc}, which recovers meaningful neighborhoods in a graph. Specifically, the diffusion matrix is defined as $\boldsymbol{S}=\sum_{r=0}^{\infty} \theta_{r} \boldsymbol{T}^{r}$, which has two popular versions of Personalized PageRank (PPR) with $\boldsymbol{T}=\boldsymbol{A}\boldsymbol{D}^{-1}$, $\theta_r=\alpha(1-\alpha)^r$, and Heat Kernel (HK) with $\boldsymbol{T}=\boldsymbol{A}\boldsymbol{D}^{-1}$, $\theta_r=e^{-t}\frac{t^r}{r!}$, where $\boldsymbol{D}$ is the diagonal matrix of node degrees, i.e., $\boldsymbol{D}_{ii}=\sum_j\boldsymbol{A}_{ij}$, and $t$ is the diffusion time. After sparsifying $\boldsymbol{S}$ as in~\cite{gdc}, we get a weighted and sparse graph adjacency matrix $\boldsymbol{\tilde{S}}\in \mathbb{R}^{N\times N}$, which can be regarded as a weighted version of the adjacency matrix $\boldsymbol{A}$.

Afterward, we can leverage $\boldsymbol{\tilde{S}}_{anc}$ as the probability of a multinomial distribution to sample neighbors of $v_{syn}$.
The number of neighbors is sampled from another degree distribution based on the entire graph to keep degree statistics, as suggested in~\cite{graphens}.

\begin{algorithm}[t]
\caption{GraphSHA synthesis algorithm}
\label{algo:graphsha}
\flushleft{\textbf{Input}: graph $\mathcal{G}=(\boldsymbol{X},\boldsymbol{A})$, 
training set nodes $\mathcal{V}^L$ and their labels $\boldsymbol{Y}^L$, 
number of classes $C$}

\textbf{Parameters}: distribution $\mathcal{D}$ to sample $\delta$ \\

\begin{algorithmic}[1] 
\STATE Initialize GNN $f_{\theta}$
\STATE Calculate $\boldsymbol{\tilde{S}}$ via graph diffusion and sparsification
\STATE Calculate degree distribution $P_{degree}$ for $\mathcal{G}$
\STATE Calculate the number of samples to synthesize $n_c$ for each class $c\in C$ 
\WHILE{not converge}
\STATE Calculate node hardness $\mathcal{H}$ for nodes in $\mathcal{V}^L$ via Eq.~\eqref{eq:hardness}
\STATE Sample anchor nodes $v_{anc}$ according to node hardness $\mathcal{H}$
\STATE Sample neighbor classes for anchor nodes
\STATE Sample auxiliary nodes $v_{aux}$ from instances in neighbor classes for anchor nodes
\STATE Sample $\delta$ from $\mathcal{D}$
\STATE Mix each $(\boldsymbol{X}_{anc},\boldsymbol{X}_{aux})$ pair via Eq.~\eqref{eq:mix} to get $\sum_c n_c$ node features $\boldsymbol{X}_{syn}$
\STATE Sample the number of edges $N^{nei}_{syn}$ from $P_{degree}$ for $v_{syn}$
\STATE Sample $N^{nei}_{syn}$ nodes from the weighted subgraph $\boldsymbol{\tilde{S}}_{anc}$ as the neighbor set for $v_{syn}$
\STATE Conduct GNN node classification on the augmented graph with synthesized nodes $v_{syn}$ and the corresponding edges
\ENDWHILE
\STATE \textbf{return} $f_{\theta}$
\end{algorithmic}
\end{algorithm}

To sum up, the feature of the synthesized minor sample $v_{syn}$ is generated via the mixup of the features of $v_{anc}$ and $v_{aux}$, while the edges connecting $v_{syn}$ is generated from the 1-hop subgraph of $v_{anc}$ without mixup. 
We give a formal definition of the synthesized harder minor sample derived from \textsc{SemiMixup} below.

\begin{definition}[harder minor sample generated by \textsc{SemiMixup}]
For nodes $v_{anc}$ and $v_{aux}$ described in Section~\ref{sec:identifying}, the synthesized harder minor sample is defined as
\begin{equation}\label{eq:semihardsample}
\left\{
\begin{aligned}
&\boldsymbol{X}_{syn}=\delta\boldsymbol{X}_{anc}+(1-\delta)\boldsymbol{X}_{aux},\\
&\mathcal{N}_{syn}\sim P_{1hop}^{diff}(v_{anc}),\\
&\boldsymbol{Y}(v_{syn})=\boldsymbol{Y}(v_{anc}),
\end{aligned}
\right.
\end{equation}
where $\delta$ is a random variable in $[0,1]$, and 
$P_{1hop}^{diff}(v_{anc})$ is the 1-hop neighbor distribution of $v_{anc}$ with probability $\boldsymbol{\tilde{S}}_{anc}$ generated via graph diffusion-based smoothing~\cite{gdc}.
\end{definition}

\subsection{Complexity Analysis}

It is worth noting that the extra calculation of GraphSHA introduces light computational overhead over the base GNN model.
Let $N_{tr}$ be the number of training samples. Then, the number of synthesized nodes for each class is $\mathcal{O}(N_{tr})$.
As logits can be obtained from the previous epoch, we can get node hardness without extra computation~\cite{conf}. 
We need $\mathcal{O}(N_{tr}^2)$ extra time for sampling anchor nodes and auxiliary nodes.
For generating node features of synthesized nodes, we need $\mathcal{O}(N_{tr} d)$ extra time, where $d$ is the dimension of the raw feature. For generating edges of synthesized nodes, we need $\mathcal{O}(N_{tr}/N \cdot |\mathcal{E}|)$ time. Overall, the extra computational overhead over the base model is $\mathcal{O}(N_{tr}^2+N_{tr}d+N_{tr}/N \cdot |\mathcal{E}|)$. As $N_{tr}\ll N$ in traditional semi-supervised node classification, GraphSHA only introduces lightweight computational overhead, which enables it to be applied to large-scale graphs, as introduced in Section~\ref{sec:large}.

\section{Experiments}\label{sec:exp}

In this section, we conduct extensive experiments to evaluate the effectiveness of GraphSHA for class-imbalanced node classification by answering the following research questions:
\begin{itemize}
\item[\textbf{RQ1:}] Does GraphSHA outperform existing baseline methods on class-imbalanced node classification?
\item[\textbf{RQ2:}] Does GraphSHA solve the squeezed minority problem? Does it actually enlarge the squeezed minor class boundary?
\item[\textbf{RQ3:}] Does the \textsc{SemiMixup} module avoid deteriorating the subspaces of major classes in the latent space?
\end{itemize}

\begin{table}[!t]
\centering
\begin{center}
\caption{Statistics of datasets used in the paper.}\label{table:datasets}
\vspace{-0.2cm}
\scalebox{0.9}{
\begin{tabular}{l|ccccc}
\toprule
{Dataset}  & {\#Nodes} & {\#Edges} & {\#Features} & {\#Classes} \\
\midrule
Cora                 & 2,708        & 10,556          & 1,433               & 7 \\
CiteSeer             & 3,327        & 9,104          & 3,703               & 6 \\
PubMed               & 19,717       & 88,648         & 500                 & 3 \\
Photo  & 7,650 & 119,081 & 745 & 8 \\
Computer  & 13,752 & 245,861 & 767 & 10 \\
CS  & 18,333 & 81,894 & 6,805 & 15 \\
arXiv & 169,343 & 1,166,243 & 128 & 40 \\
\bottomrule
\end{tabular}
}
\end{center}
\vspace{-0.2cm}
\end{table}

\begin{table*}[!t]
\centering
\begin{center}
\caption{Node classification results ($\pm$std) on Cora, CiteSeer, and PubMed in long-tailed class-imbalanced setting with GCN, GAT, and GraphSAGE backbones for 10 runs. The best is highlighted in \textbf{boldface}, and the runner-up is highlighted in \underline{underline}.}\label{table:lt}
\vspace{-0.1cm}
\scalebox{0.99}{
\begin{tabular}{cl|ccc|ccc|ccc}
\toprule
\multirow{2}{*}{} & \textbf{Dataset} & \multicolumn{3}{c|}{Cora-LT} & \multicolumn{3}{c|}{CiteSeer-LT} & \multicolumn{3}{c}{PubMed-LT} \\
\cmidrule(lr){2-2}\cmidrule(lr){3-5}\cmidrule(lr){6-8}\cmidrule(lr){9-11}
& $\rho$=100 & {Acc.} & {bAcc.} & {F1} & {Acc.} & {bAcc.} & {F1} & {Acc.} & {bAcc.} & {F1} \\
\midrule
\midrule
\multirow{11}{*}{\rotatebox{90}{GCN}}
& Vanilla & 72.02\footnotesize{$\pm$0.50} & 59.42\footnotesize{$\pm$0.74} & 59.23\footnotesize{$\pm$1.02} & 51.40\footnotesize{$\pm$0.44} & 44.64\footnotesize{$\pm$0.42} & 37.82\footnotesize{$\pm$0.67} & 51.58\footnotesize{$\pm$0.60} & 42.11\footnotesize{$\pm$0.48} & 34.73\footnotesize{$\pm$0.71} \\
\cmidrule(lr){2-11}
& Reweight & 78.42\footnotesize{$\pm$0.10} & 72.66\footnotesize{$\pm$0.17} & 73.75\footnotesize{$\pm$0.15} & \underline{63.61}\footnotesize{$\pm$0.22} & 56.80\footnotesize{$\pm$0.20} & 55.18\footnotesize{$\pm$0.18} & 77.02\footnotesize{$\pm$0.14} & 72.45\footnotesize{$\pm$0.17} & 72.12\footnotesize{$\pm$0.15} \\
& PC Softmax & 77.30\footnotesize{$\pm$0.13} & 72.08\footnotesize{$\pm$0.30} & 71.65\footnotesize{$\pm$0.34} & 62.15\footnotesize{$\pm$0.45} & \textbf{59.08}\footnotesize{$\pm$0.28} & \underline{58.13}\footnotesize{$\pm$0.31} & 74.36\footnotesize{$\pm$0.62} & 72.59\footnotesize{$\pm$0.34} & 71.79\footnotesize{$\pm$0.50} \\
& CB Loss & 77.97\footnotesize{$\pm$0.19} & 72.70\footnotesize{$\pm$0.28} & 73.17\footnotesize{$\pm$0.22} & 61.47\footnotesize{$\pm$0.51} & 55.18\footnotesize{$\pm$0.52} & 53.47\footnotesize{$\pm$0.65} & 76.57\footnotesize{$\pm$0.19} & 72.16\footnotesize{$\pm$0.18} & 72.84\footnotesize{$\pm$0.19} \\
& Focal Loss & 78.43\footnotesize{$\pm$0.19} & \underline{73.17}\footnotesize{$\pm$0.23} & 73.76\footnotesize{$\pm$0.20} & 59.66\footnotesize{$\pm$0.38} & 53.39\footnotesize{$\pm$0.33} & 51.80\footnotesize{$\pm$0.39} & 75.67\footnotesize{$\pm$0.20} & 71.34\footnotesize{$\pm$0.24} & 72.03\footnotesize{$\pm$0.21} \\
& ReNode & \underline{78.93}\footnotesize{$\pm$0.13} & 73.13\footnotesize{$\pm$0.17} & \underline{74.46}\footnotesize{$\pm$0.16} & 62.39\footnotesize{$\pm$0.31} & 55.62\footnotesize{$\pm$0.27} & 54.05\footnotesize{$\pm$0.24} & 76.00\footnotesize{$\pm$0.16} & 70.68\footnotesize{$\pm$0.15} & 71.41\footnotesize{$\pm$0.15} \\
\cmidrule(lr){2-11}
& Upsample & 75.52\footnotesize{$\pm$0.11} & 66.68\footnotesize{$\pm$0.14} & 68.35\footnotesize{$\pm$0.15} & 55.05\footnotesize{$\pm$0.11} & 48.41\footnotesize{$\pm$0.11} & 45.22\footnotesize{$\pm$0.14} & 71.58\footnotesize{$\pm$0.06} & 63.79\footnotesize{$\pm$0.06} & 64.62\footnotesize{$\pm$0.07} \\
& GraphSmote & 75.44\footnotesize{$\pm$0.43} & 68.99\footnotesize{$\pm$0.51} & 70.41\footnotesize{$\pm$0.52} & 56.58\footnotesize{$\pm$0.29 }& 50.39\footnotesize{$\pm$0.28} & 47.96\footnotesize{$\pm$0.33} & 74.62\footnotesize{$\pm$0.08} & 69.53\footnotesize{$\pm$0.10} & 71.18\footnotesize{$\pm$0.09} \\
& GraphENS & 76.15\footnotesize{$\pm$0.24} & 71.16\footnotesize{$\pm$0.40} & 70.85\footnotesize{$\pm$0.49} & 63.14\footnotesize{$\pm$0.35} & 56.92\footnotesize{$\pm$0.37} & 55.54\footnotesize{$\pm$0.41} & 77.11\footnotesize{$\pm$0.11} & 71.89\footnotesize{$\pm$0.15} & 72.71\footnotesize{$\pm$0.14} \\
& TAM (G-ENS) & 77.30\footnotesize{$\pm$0.23} & 72.10\footnotesize{$\pm$0.29} & 72.25\footnotesize{$\pm$0.29} & 63.40\footnotesize{$\pm$0.34} & 57.15\footnotesize{$\pm$0.35} & 55.68\footnotesize{$\pm$0.40} & \underline{78.07}\footnotesize{$\pm$0.15} & \underline{72.63}\footnotesize{$\pm$0.23} & \underline{72.96}\footnotesize{$\pm$0.22} \\
& \cellcolor{light-gray}\textbf{GraphSHA} & \cellcolor{light-gray}\textbf{79.90}\footnotesize{$\pm$0.29} & \cellcolor{light-gray}\textbf{74.62}\footnotesize{$\pm$0.35} & \cellcolor{light-gray}\textbf{75.74}\footnotesize{$\pm$0.32} & \cellcolor{light-gray}\textbf{64.50}\footnotesize{$\pm$0.41} & \cellcolor{light-gray}\underline{59.04}\footnotesize{$\pm$0.34} & \cellcolor{light-gray}\textbf{59.16}\footnotesize{$\pm$0.21} & \cellcolor{light-gray}\textbf{79.20}\footnotesize{$\pm$0.13} & \cellcolor{light-gray}\textbf{74.46}\footnotesize{$\pm$0.17} & \cellcolor{light-gray}\textbf{75.24}\footnotesize{$\pm$0.27} \\
\midrule
\midrule
\multirow{11}{*}{\rotatebox{90}{GAT}}
& Vanilla & 67.52\footnotesize{$\pm$0.58} & 54.20\footnotesize{$\pm$0.79} & 55.34\footnotesize{$\pm$0.74} & 49.16\footnotesize{$\pm$0.19} & 42.58\footnotesize{$\pm$0.18} & 35.75\footnotesize{$\pm$0.29} & 47.83\footnotesize{$\pm$1.57} & 39.09\footnotesize{$\pm$1.27} & 29.62\footnotesize{$\pm$2.15} \\
\cmidrule(lr){2-11}
& Reweight & 77.77\footnotesize{$\pm$0.28} & 72.03\footnotesize{$\pm$0.58} & 72.79\footnotesize{$\pm$0.58} & 61.95\footnotesize{$\pm$0.57} & 55.40\footnotesize{$\pm$0.63} & 53.71\footnotesize{$\pm$0.72} & 74.08\footnotesize{$\pm$0.39} & 69.35\footnotesize{$\pm$0.79} & 69.52\footnotesize{$\pm$0.70} \\
& PC Softmax & 68.75\footnotesize{$\pm$0.77} & 64.07\footnotesize{$\pm$0.86} & 64.17\footnotesize{$\pm$0.74} & 56.70\footnotesize{$\pm$1.61} & 56.31\footnotesize{$\pm$1.30} & 55.31\footnotesize{$\pm$1.53} & 76.70\footnotesize{$\pm$0.32} & \underline{73.22}\footnotesize{$\pm$0.15} & \underline{73.25}\footnotesize{$\pm$0.28} \\
& CB Loss & 77.29\footnotesize{$\pm$0.36} & 72.07\footnotesize{$\pm$0.63} & 72.79\footnotesize{$\pm$0.43} & 61.44\footnotesize{$\pm$0.52} & 55.17\footnotesize{$\pm$0.52} & 53.63\footnotesize{$\pm$0.54} & 74.81\footnotesize{$\pm$0.25} & 69.54\footnotesize{$\pm$0.64} & 70.55\footnotesize{$\pm$0.59} \\
& Focal Loss & 77.97\footnotesize{$\pm$0.11} & 72.47\footnotesize{$\pm$0.21} & 73.15\footnotesize{$\pm$0.21} & 59.75\footnotesize{$\pm$0.36} & 53.44\footnotesize{$\pm$0.34} & 52.12\footnotesize{$\pm$0.29} & 74.23\footnotesize{$\pm$0.27} & 70.36\footnotesize{$\pm$0.34} & 70.63\footnotesize{$\pm$0.20} \\
& ReNode & \underline{78.09}\footnotesize{$\pm$0.24} & 71.78\footnotesize{$\pm$0.34} & \underline{73.41}\footnotesize{$\pm$0.34} & 60.87\footnotesize{$\pm$0.37} & 54.01\footnotesize{$\pm$0.37} & 51.98\footnotesize{$\pm$0.44} & 74.09\footnotesize{$\pm$0.28} & 69.02\footnotesize{$\pm$0.34} & 69.55\footnotesize{$\pm$0.32} \\
\cmidrule(lr){2-11}
& Upsample & 72.62\footnotesize{$\pm$0.31} & 62.39\footnotesize{$\pm$0.37} & 65.08\footnotesize{$\pm$0.28} & 53.41\footnotesize{$\pm$0.22} & 46.89\footnotesize{$\pm$0.22} & 43.10\footnotesize{$\pm$0.44} & 67.61\footnotesize{$\pm$0.95} & 57.29\footnotesize{$\pm$0.64} & 54.99\footnotesize{$\pm$1.02} \\
& GraphSmote & 74.65\footnotesize{$\pm$0.29} & 67.71\footnotesize{$\pm$0.37} & 69.10\footnotesize{$\pm$0.39} & 57.45\footnotesize{$\pm$0.26} & 51.33\footnotesize{$\pm$0.31} & 49.38\footnotesize{$\pm$0.54} & 74.04\footnotesize{$\pm$0.38} & 69.04\footnotesize{$\pm$0.35} & 70.62\footnotesize{$\pm$0.42} \\
& GraphENS & 77.08\footnotesize{$\pm$0.26} & 72.07\footnotesize{$\pm$0.38} & 72.09\footnotesize{$\pm$0.48} & 61.91\footnotesize{$\pm$0.34} & 55.88\footnotesize{$\pm$0.32} & 54.38\footnotesize{$\pm$0.41} & 76.65\footnotesize{$\pm$0.11} & 70.43\footnotesize{$\pm$0.20} & 71.25\footnotesize{$\pm$0.20} \\
& TAM (G-ENS) & 77.69\footnotesize{$\pm$0.21} & \underline{72.87}\footnotesize{$\pm$0.30} & 72.99\footnotesize{$\pm$0.31} & \textbf{64.06}\footnotesize{$\pm$0.34} & \underline{57.77}\footnotesize{$\pm$0.31} & \underline{56.38}\footnotesize{$\pm$0.32} & \underline{77.94}\footnotesize{$\pm$0.18} & 71.98\footnotesize{$\pm$0.29} & 73.07\footnotesize{$\pm$0.27} \\
& \cellcolor{light-gray}\textbf{GraphSHA} & \cellcolor{light-gray}\textbf{79.07}\footnotesize{$\pm$0.18} & \cellcolor{light-gray}\textbf{74.08}\footnotesize{$\pm$0.26} & \cellcolor{light-gray}\textbf{75.02}\footnotesize{$\pm$0.18} & \cellcolor{light-gray}\underline{63.94}\footnotesize{$\pm$0.44} & \cellcolor{light-gray}\textbf{58.14}\footnotesize{$\pm$0.35} & \cellcolor{light-gray}\textbf{57.71}\footnotesize{$\pm$0.40} & \cellcolor{light-gray}\textbf{78.40}\footnotesize{$\pm$0.20} & \cellcolor{light-gray}\textbf{73.82}\footnotesize{$\pm$0.17} & \cellcolor{light-gray}\textbf{74.66}\footnotesize{$\pm$0.21} \\
\midrule
\midrule
\multirow{11}{*}{\rotatebox{90}{SAGE}}
& Vanilla & 73.30\footnotesize{$\pm$0.09} & 61.83\footnotesize{$\pm$0.12} & 63.25\footnotesize{$\pm$0.13} & 47.90\footnotesize{$\pm$0.24} & 41.80\footnotesize{$\pm$0.22} & 36.96\footnotesize{$\pm$0.31} & 58.78\footnotesize{$\pm$0.08} & 47.92\footnotesize{$\pm$0.06} & 42.34\footnotesize{$\pm$0.07} \\
\cmidrule(lr){2-11}
& Reweight & 76.81\footnotesize{$\pm$0.15} & 68.74\footnotesize{$\pm$0.31} & 70.22\footnotesize{$\pm$0.37} & 57.30\footnotesize{$\pm$0.53} & 50.90\footnotesize{$\pm$0.46} & 49.15\footnotesize{$\pm$0.49} & 65.94\footnotesize{$\pm$0.53} & 59.83\footnotesize{$\pm$1.24} & 58.89\footnotesize{$\pm$1.14} \\
& PC Softmax & 76.92\footnotesize{$\pm$0.22} & \underline{73.25}\footnotesize{$\pm$0.28} & \underline{73.54}\footnotesize{$\pm$0.26} & 58.35\footnotesize{$\pm$0.25} & 56.06\footnotesize{$\pm$0.18} & \underline{56.65}\footnotesize{$\pm$0.18} & 71.60\footnotesize{$\pm$0.13} & \underline{73.83}\footnotesize{$\pm$0.15} & 70.28\footnotesize{$\pm$0.12} \\
& CB Loss & 77.04\footnotesize{$\pm$0.30} & 70.25\footnotesize{$\pm$0.37} & 71.26\footnotesize{$\pm$0.30} & 57.63\footnotesize{$\pm$0.34} & 51.19\footnotesize{$\pm$0.32} & 48.70\footnotesize{$\pm$0.35} & 67.78\footnotesize{$\pm$0.36} & 60.67\footnotesize{$\pm$0.46} & 61.46\footnotesize{$\pm$0.52} \\
& Focal Loss & 77.17\footnotesize{$\pm$0.16} & 69.78\footnotesize{$\pm$0.27} & 70.76\footnotesize{$\pm$0.25} & 57.02\footnotesize{$\pm$0.72} & 50.77\footnotesize{$\pm$0.66} & 48.42\footnotesize{$\pm$0.79} & 70.59\footnotesize{$\pm$0.35} & 65.69\footnotesize{$\pm$0.45} & 66.25\footnotesize{$\pm$0.44} \\
& ReNode & 77.26\footnotesize{$\pm$0.15} & 69.22\footnotesize{$\pm$0.21} & 71.13\footnotesize{$\pm$0.22} & 57.82\footnotesize{$\pm$0.50} & 51.27\footnotesize{$\pm$0.49} & 49.04\footnotesize{$\pm$0.45} & 67.60\footnotesize{$\pm$0.51} & 60.65\footnotesize{$\pm$0.82} & 60.78\footnotesize{$\pm$0.78} \\
\cmidrule(lr){2-11}
& Upsample & 73.80\footnotesize{$\pm$0.12} & 63.45\footnotesize{$\pm$0.20} & 65.83\footnotesize{$\pm$0.16} & 50.32\footnotesize{$\pm$0.11} & 44.24\footnotesize{$\pm$0.11} & 41.46\footnotesize{$\pm$0.17} & 64.08\footnotesize{$\pm$0.06} & 54.64\footnotesize{$\pm$0.07} & 53.39\footnotesize{$\pm$0.10} \\
& GraphSmote & 74.24\footnotesize{$\pm$0.19} & 66.15\footnotesize{$\pm$0.38} & 67.89\footnotesize{$\pm$0.41} & 52.85\footnotesize{$\pm$0.64} & 46.99\footnotesize{$\pm$0.63} & 44.20\footnotesize{$\pm$0.74} & 65.10\footnotesize{$\pm$0.42} & 56.82\footnotesize{$\pm$0.49} & 56.85\footnotesize{$\pm$0.54} \\
& GraphENS & 76.69\footnotesize{$\pm$0.20} & 70.07\footnotesize{$\pm$0.25} & 70.37\footnotesize{$\pm$0.30} &62.63\footnotesize{$\pm$0.34} & 56.14\footnotesize{$\pm$0.37} & 54.13\footnotesize{$\pm$0.39} & 77.62\footnotesize{$\pm$0.14} & 72.54\footnotesize{$\pm$0.23} & 73.21\footnotesize{$\pm$0.18} \\
& TAM (G-ENS) & \underline{77.31}\footnotesize{$\pm$0.30} & 71.02\footnotesize{$\pm$0.34} & 71.14\footnotesize{$\pm$0.36} & \underline{62.93}\footnotesize{$\pm$0.21} & \underline{56.44}\footnotesize{$\pm$0.19} & 54.50\footnotesize{$\pm$0.21} &\underline{78.12}\footnotesize{$\pm$0.31} & 72.80\footnotesize{$\pm$0.76} & \underline{73.69}\footnotesize{$\pm$0.67} \\
& \cellcolor{light-gray}\textbf{GraphSHA} & \cellcolor{light-gray}\textbf{78.80}\footnotesize{$\pm$0.24} & \cellcolor{light-gray}\textbf{73.56}\footnotesize{$\pm$0.35} & \cellcolor{light-gray}\textbf{74.27}\footnotesize{$\pm$0.30} & \cellcolor{light-gray}\textbf{63.76}\footnotesize{$\pm$0.38} & \cellcolor{light-gray}\textbf{58.25}\footnotesize{$\pm$0.37} & \cellcolor{light-gray}\textbf{58.04}\footnotesize{$\pm$0.45} & \cellcolor{light-gray}\textbf{78.20}\footnotesize{$\pm$0.19} & \cellcolor{light-gray}\textbf{74.07}\footnotesize{$\pm$0.23} & \cellcolor{light-gray}\textbf{74.93}\footnotesize{$\pm$0.23} \\
\bottomrule
\end{tabular}
}
\end{center}
\vspace{-0.2cm}
\end{table*}

\begin{table*}[!t]
\centering
\begin{center}
\caption{Node classification results ($\pm$std) on Photo, Computer, and CS in step class-imbalanced setting with GraphSAGE.}\label{table:step}
\vspace{-0.1cm}
\scalebox{0.99}{
\begin{tabular}{cl|ccc|ccc|ccc}
\toprule
\multirow{2}{*}{} & \textbf{Dataset} & \multicolumn{3}{c|}{Photo-ST} & \multicolumn{3}{c|}{Computer-ST} & \multicolumn{3}{c}{CS-ST} \\
\cmidrule(lr){2-2}\cmidrule(lr){3-5}\cmidrule(lr){6-8}\cmidrule(lr){9-11}
& $\rho$=20 & {Acc.} & {bAcc.} & {F1} & {Acc.} & {bAcc.} & {F1} & {Acc.} & {bAcc.} & {F1} \\
\midrule
\midrule
\multirow{11}{*}{\rotatebox{90}{SAGE}}
& Vanilla & 59.19\footnotesize{$\pm$2.32} & 59.98\footnotesize{$\pm$1.82} & 47.11\footnotesize{$\pm$2.95} & 63.88\footnotesize{$\pm$0.05} & 46.96\footnotesize{$\pm$0.04} & 30.08\footnotesize{$\pm$0.11} & 74.81\footnotesize{$\pm$0.35} & 79.69\footnotesize{$\pm$0.19} & 64.68\footnotesize{$\pm$0.52} \\
\cmidrule(lr){2-11}
& Reweight & 84.85\footnotesize{$\pm$0.23} & 87.30\footnotesize{$\pm$0.24} & 82.89\footnotesize{$\pm$0.16} & 83.59\footnotesize{$\pm$0.31} & 87.91\footnotesize{$\pm$0.10} & \underline{77.59}\footnotesize{$\pm$0.41} & 91.02\footnotesize{$\pm$0.33} & 90.87\footnotesize{$\pm$0.30} & 75.50\footnotesize{$\pm$0.35} \\
& PC Softmax & 86.16\footnotesize{$\pm$0.13} & 86.93\footnotesize{$\pm$0.15} & 83.55\footnotesize{$\pm$0.09} & 81.38\footnotesize{$\pm$0.17} & 80.50\footnotesize{$\pm$0.76} & 72.30\footnotesize{$\pm$0.57} & 92.58\footnotesize{$\pm$0.23} & 92.11\footnotesize{$\pm$0.37} & 78.00\footnotesize{$\pm$0.53} \\
& CB Loss & 83.02\footnotesize{$\pm$0.29} & 85.79\footnotesize{$\pm$0.21} & 80.48\footnotesize{$\pm$0.31} & \underline{83.75}\footnotesize{$\pm$0.21} & 87.38\footnotesize{$\pm$0.10} & 77.08\footnotesize{$\pm$0.20} & 90.85\footnotesize{$\pm$0.14} & 90.77\footnotesize{$\pm$0.10} & 79.04\footnotesize{$\pm$0.85} \\
& Focal Loss & 82.58\footnotesize{$\pm$0.39} & 85.42\footnotesize{$\pm$0.32} & 79.28\footnotesize{$\pm$0.35} & 82.56\footnotesize{$\pm$0.22} & 87.38\footnotesize{$\pm$0.08} & 76.53\footnotesize{$\pm$0.15} & 90.08\footnotesize{$\pm$0.19} & 90.01\footnotesize{$\pm$0.16} & 79.56\footnotesize{$\pm$0.26} \\
& ReNode & 84.83\footnotesize{$\pm$0.15} & 86.43\footnotesize{$\pm$0.20} & 81.85\footnotesize{$\pm$0.22} & 81.29\footnotesize{$\pm$0.34} & 87.33\footnotesize{$\pm$0.17} & 76.60\footnotesize{$\pm$0.28} & 90.98\footnotesize{$\pm$0.31} & 91.17\footnotesize{$\pm$0.35} & \textbf{81.22}\footnotesize{$\pm$0.43} \\
\cmidrule(lr){2-11}
	& Upsample & 82.20\footnotesize{$\pm$0.34} & 84.86\footnotesize{$\pm$0.08} & 79.38\footnotesize{$\pm$0.24} & 82.99\footnotesize{$\pm$0.24} & 87.02\footnotesize{$\pm$0.09} & 77.10\footnotesize{$\pm$0.33} & 87.23\footnotesize{$\pm$0.18} & 87.99\footnotesize{$\pm$0.11} & 76.38\footnotesize{$\pm$0.21} \\
& GraphSmote & 80.21\footnotesize{$\pm$0.27} & 84.68\footnotesize{$\pm$0.31} & 79.05\footnotesize{$\pm$0.38} & 83.62\footnotesize{$\pm$0.25} & 88.15\footnotesize{$\pm$0.21} & 76.02\footnotesize{$\pm$0.30} & 86.30\footnotesize{$\pm$0.12} & 85.66\footnotesize{$\pm$0.09} & 69.19\footnotesize{$\pm$0.14} \\
& GraphENS & \underline{88.02}\footnotesize{$\pm$0.09} & \underline{90.55}\footnotesize{$\pm$0.11} & \underline{86.70}\footnotesize{$\pm$0.10} & 83.28\footnotesize{$\pm$0.38} & \underline{88.54}\footnotesize{$\pm$0.10} & 76.77\footnotesize{$\pm$0.52} & 92.13\footnotesize{$\pm$0.16} & \underline{92.53}\footnotesize{$\pm$0.22} & 78.23\footnotesize{$\pm$0.20} \\
& TAM (G-ENS) & 87.61\footnotesize{$\pm$0.13} & 89.17\footnotesize{$\pm$0.17} & 85.74\footnotesize{$\pm$0.16} & 80.31\footnotesize{$\pm$0.52} & 86.74\footnotesize{$\pm$0.22} & 76.96\footnotesize{$\pm$0.59} & \underline{92.60}\footnotesize{$\pm$0.22} & 92.39\footnotesize{$\pm$0.19} & 78.52\footnotesize{$\pm$0.23} \\
& \cellcolor{light-gray}\textbf{GraphSHA} & \cellcolor{light-gray}\textbf{89.14}\footnotesize{$\pm$0.22} & \cellcolor{light-gray}\textbf{90.60}\footnotesize{$\pm$0.10} & \cellcolor{light-gray}\textbf{87.25}\footnotesize{$\pm$0.18} & \cellcolor{light-gray}\textbf{84.21}\footnotesize{$\pm$0.50} & \cellcolor{light-gray}\textbf{89.49}\footnotesize{$\pm$0.10} & \cellcolor{light-gray}\textbf{77.93}\footnotesize{$\pm$0.91} & \cellcolor{light-gray}\textbf{92.93}\footnotesize{$\pm$0.05} & \cellcolor{light-gray}\textbf{92.78}\footnotesize{$\pm$0.13} & \cellcolor{light-gray}\underline{79.68}\footnotesize{$\pm$0.16} \\
\bottomrule
\end{tabular}
}
\end{center}
\vspace{-0.1cm}
\end{table*}

\vspace{-0.1cm}
\subsection{Experimental Setup}

\subsubsection*{\textbf{Datasets}} 
We adopt seven benchmark datasets, including Cora, CiteSeer, PubMed~\cite{cora}, Amazon-Photo (Photo), Amazon-Computers (computer), Coauthor-CS (CS)~\cite{amazon} and ogbn-arXiv (arXiv)~\cite{ogb} to conduct all the experiments. The statistics of these datasets are provided in Table~\ref{table:datasets}. A detailed description of them is provided in Appendix~\ref{appendix:datasets}.

\subsubsection*{\textbf{Compared Baselines}} 
We compare GraphSHA with various imbalance handling methods. 
For loss-modifying approaches, we compare (1) Reweight, which reweights class weights to be proportional to the numbers of class samples; (2) PC Softmax~\cite{pcs} and (3) Class-Balanced Loss (CB Loss)~\cite{cb}, which are two general methods to modify the loss function in handling the imbalance issue; (4) Focal Loss~\cite{focal}, which modifies the loss function to focus on hard samples; and (5) ReNode~\cite{renode}, which also considers topological imbalance in the context of graph data based on the class imbalance. 
For generative approaches, we compare (1) Upsample, which directly duplicates minor nodes; (2) GraphSmote~\cite{graphsmote} and (3) GraphENS~\cite{graphens}, which have been introduced in Section~\ref{sec:related}. Furthermore, we also compare the state-of-the-art method (4) TAM~\cite{tam} based on the best-performed GraphENS by default, which aims to decrease the false positive cases considering the graph topology.

\vspace{-0.3cm}
\subsubsection*{\textbf{Configurations and Evaluation Protocols}} 
We adopt various GNNs, namely GCN~\cite{gcn}, GAT~\cite{gat}, and GraphSAGE~\cite{graphsage} as the backbones of our model and baseline models. Their hidden layer is set to 2, both in 64-dim by default. For GAT, the multi-head number is set to 8. For our GraphSHA, $\delta$ for node feature synthesis is sampled from a beta distribution as $\delta\sim beta(b_1,b_2)$, and we present the hyper-parameter analysis of the distribution in Section~\ref{sec:hyper}.
For diffusion matrix $\boldsymbol{S}$, we adopt the PPR version with $\alpha=0.05$, and we also adopt top-$K$ with $K=128$ to select $K$ highest mass per column of $\boldsymbol{S}$ to get a sparsified $\boldsymbol{\tilde{S}}$, both of which are suggested in~\cite{gdc}.
The implementation details for baselines are described in Appendix~\ref{appendix:baselines}.
Samples are synthesized until each class reaches the mean or maximum number of samples among all classes in the training set as a hyper-parameter.
We apply Accuracy (Acc.), balanced Accuracy (bAcc.), and macro F1 score (F1) as the evaluation metrics following~\cite{graphens, tam}, where bAcc. is defined as the average recall for each class~\cite{bacc}.

\vspace{-0.3cm}
\subsection{Results on Manually Imbalanced Datasets}
We consider both long-tailed class imbalance setting~\cite{graphens} on Cora, CiteSeer, PubMed and step class imbalance setting~\cite{graphsmote, renode} on Photo, Computer, and CS to conduct the experiments.
In the long-tailed setting, we adopt full data split~\cite{corafull} for the three datasets, and we remove labeled nodes in the training set manually until they follow a long-tailed distribution as in~\cite{cb, graphens}.
The imbalance ratio $\rho$ is set to an extreme condition of 100. In step setting, the datasets are split into training/validation/test sets with proportions 10\%/10\%/80\% respectively as in~\cite{graphsmote, renode}, where half of the classes are major classes and share the same number of training samples $n_{maj}$, while the other half are minor classes and share the same number of training samples $n_{min}=n_{maj}/ \rho$ in the training set. The imbalance ratio $\rho$ is set to 20 in this setting. The results are shown in Table~\ref{table:lt} and Table~\ref{table:step} respectively for the two settings.

From both tables, we can find that GraphSHA shows significant improvements compared with almost all other contenders with different GNN backbones, which shows the effectiveness of the overall framework of GraphSHA. We provide more observations as follows. Firstly, the performances of different methods are similar across different GNN backbones, which shows that the performance gaps result from the models' intrinsic properties. Secondly, generative approaches generally perform better than loss-modifying approaches, which benefits from the augmented topological structure. The results in step setting with GCN and GAT backbones are provided in Appendix~\ref{appendix:exp} due to the space constraint.

\subsection{Results on Naturally Imbalanced Datasets}\label{sec:large}

\begin{table}[!t]
\centering
\begin{center}
\caption{Node classification results ($\pm$std) on large-scale naturally class-imbalanced dataset ogbn-arXiv. OOM indicates Out-Of-Memory on a 24GB GPU.}\label{table:ogbn}
\vspace{-0.2cm}
\scalebox{0.94}{
\begin{tabular}{l|cccc}
\toprule
{\textbf{Method}} & {Val Acc.} & {Test Acc.} & {Test bAcc.} & {Test F1} \\
\midrule
Vanilla (GCN) & \underline{73.02}\footnotesize{$\pm$0.14} & \underline{71.81}\footnotesize{$\pm$0.26} & 50.96\footnotesize{$\pm$0.21} & 50.42\footnotesize{$\pm$0.18} \\
\midrule
Reweight & 67.49\footnotesize{$\pm$0.32} & 66.07\footnotesize{$\pm$0.55} & 53.34\footnotesize{$\pm$0.30} & 48.07\footnotesize{$\pm$0.77} \\
PC Softmax & 72.19\footnotesize{$\pm$0.11} & 71.49\footnotesize{$\pm$0.25} & 48.14\footnotesize{$\pm$0.14} & \underline{50.59}\footnotesize{$\pm$0.13} \\
CB Loss & 65.75\footnotesize{$\pm$0.23} & 64.73\footnotesize{$\pm$0.86} & 52.66\footnotesize{$\pm$0.72} & 47.24\footnotesize{$\pm$1.25} \\
Focal Loss & 67.36\footnotesize{$\pm$0.24} & 65.93\footnotesize{$\pm$0.58} & 53.06\footnotesize{$\pm$0.21} & 48.89\footnotesize{$\pm$0.72} \\
ReNode & 66.44\footnotesize{$\pm$0.51} & 65.91\footnotesize{$\pm$0.20} & 53.39\footnotesize{$\pm$0.40} & 48.18\footnotesize{$\pm$0.52} \\
TAM (ReNode) & 67.91\footnotesize{$\pm$0.27} & 66.63\footnotesize{$\pm$0.66} & \underline{53.40}\footnotesize{$\pm$0.24} & 48.71\footnotesize{$\pm$0.49} \\
\midrule
Upsample & 70.53\footnotesize{$\pm$0.08} & 69.55\footnotesize{$\pm$0.37} & 46.82\footnotesize{$\pm$0.07} & 45.49\footnotesize{$\pm$0.20} \\
GraphSmote & OOM & OOM & OOM & OOM \\
GraphENS & OOM & OOM & OOM & OOM \\
\rowcolor{light-gray}\textbf{GraphSHA} & \textbf{73.04}\footnotesize{$\pm$0.11} & \textbf{72.14}\footnotesize{$\pm$0.28} & \textbf{53.75}\footnotesize{$\pm$0.16} & \textbf{53.13}\footnotesize{$\pm$0.20} \\
\bottomrule
\end{tabular}
}
\end{center}
\vspace{-0.5cm}
\end{table}

Class-imbalance problem is believed to be a common issue on real-world graphs~\cite{graphsmote}, especially for those large-scale ones. We adopt arXiv dataset from OGB benchmark~\cite{ogb}, which is highly imbalanced with a training imbalance ratio of 775, validation imbalance ratio of 2,282, test imbalance ratio of 2,148, and overall imbalance ratio of 942. Though the dataset is highly imbalanced, this problem is barely studied in previous work.

\begin{table*}[!t]
\centering
\begin{center}
\caption{Ablation study of each component of GraphSHA evaluated on Cora-LT with GCN. ``+'' stands for synthesizing.}\label{table:ablation}
\scalebox{0.95}{
\begin{tabular}{l|ccc|ccccccc}
\toprule
{\textbf{Method}} & {Acc.} & {bAcc.} & {F1} & {\tabincell{c}{C0 \\ (0.5\%)}} & {\tabincell{c}{C1 \\ (1.1\%)}} & {\tabincell{c}{C2 \\ (2.4\%)}} & {\tabincell{c}{C3 \\(5.4\%)}} & {\tabincell{c}{C4 \\ (11.6\%)}} & {\tabincell{c}{C5 \\ (25.0\%)}} & {\tabincell{c}{C6 \\ (54.0\%)}} \\
\midrule
GCN & 72.02\footnotesize{$\pm$0.50} & 59.42\footnotesize{$\pm$0.74} & 59.23\footnotesize{$\pm$1.02} & 0.0 & 28.6 & 67.0 & 60.0 & 81.2 & 93.8 & 93.1 \\
\quad+easy samples & 76.90\footnotesize{$\pm$0.19} & 69.55\footnotesize{$\pm$0.21} & 71.28\footnotesize{$\pm$0.25} & 21.1 & 69.4 & 67.9 & 63.1 & 73.1 & 95.1 & 94.7 \\
\quad+harder samples w/o \textsc{SemiMixup} & 75.84\footnotesize{$\pm$0.38} & 71.38\footnotesize{$\pm$0.58} & 71.44\footnotesize{$\pm$0.59} & 54.7 & 71.7 & 63.1 & 58.4 & 74.2 & 92.5 & 82.6 \\
\midrule
\quad+harder samples w/ \textsc{SemiMixup} & 79.16\footnotesize{$\pm$0.25} & 72.89\footnotesize{$\pm$0.32} & 74.62\footnotesize{$\pm$0.27} & 42.2 & 74.3 & 71.8 & 62.3 & 72.5 & 94.4 & 93.4 \\
\quad+harder samples w/ \textsc{SemiMixup} (HK) & 79.60\footnotesize{$\pm$0.17} & 74.37\footnotesize{$\pm$0.18} & 75.17\footnotesize{$\pm$0.15} & 48.4 & 75.8 & 68.3 & 63.2 & 77.8 & 93.5 & 92.8 \\
\quad+harder samples w/ \textsc{SemiMixup} (PPR) & 79.90\footnotesize{$\pm$0.29} & 74.62\footnotesize{$\pm$0.35} & 75.74\footnotesize{$\pm$0.32} & 51.6 & 76.9 & 66.0 & 65.4 & 76.5 & 93.8 & 92.1 \\
\bottomrule
\end{tabular}
}
\end{center}
\end{table*}

The result is shown in Table~\ref{table:ogbn}. As GraphENS suffers from Out-Of-Memory, we conduct TAM based on ReNode. We report the accuracy on validation and test sets with hidden layer size 256, which is a common setting for this task as the dataset is split based on chronological order. We also report balanced accuracy and F1 score on the test set. From the table we can see that \textbf{(1)} Nearly all imbalance handling approaches can improve balance accuracy. However, accuracy and F1 score are reduced compared with vanilla GCN for the baselines, which we attribute to the decision boundary of minor classes not being properly enlarged as the boundaries of major classes are seriously degenerated. On the other hand, our GraphSHA outperforms all baselines in terms of all metrics, which verifies that it can enlarge the minor subspaces properly via the \textsc{SemiMixup} module to avoid violating neighbor classes. \textbf{(2)} Generative approaches GraphSmote and GraphENS both suffer from the OOM issue, which results from the calculation of the nearest sample in the latent space and adjacent node distribution, respectively. On the other hand, Our GraphSHA introduces light extra computational overhead by effectively choosing source nodes via the confidence-based node hardness.

\begin{figure}[!t]
\centering
\includegraphics[width=0.8\linewidth]{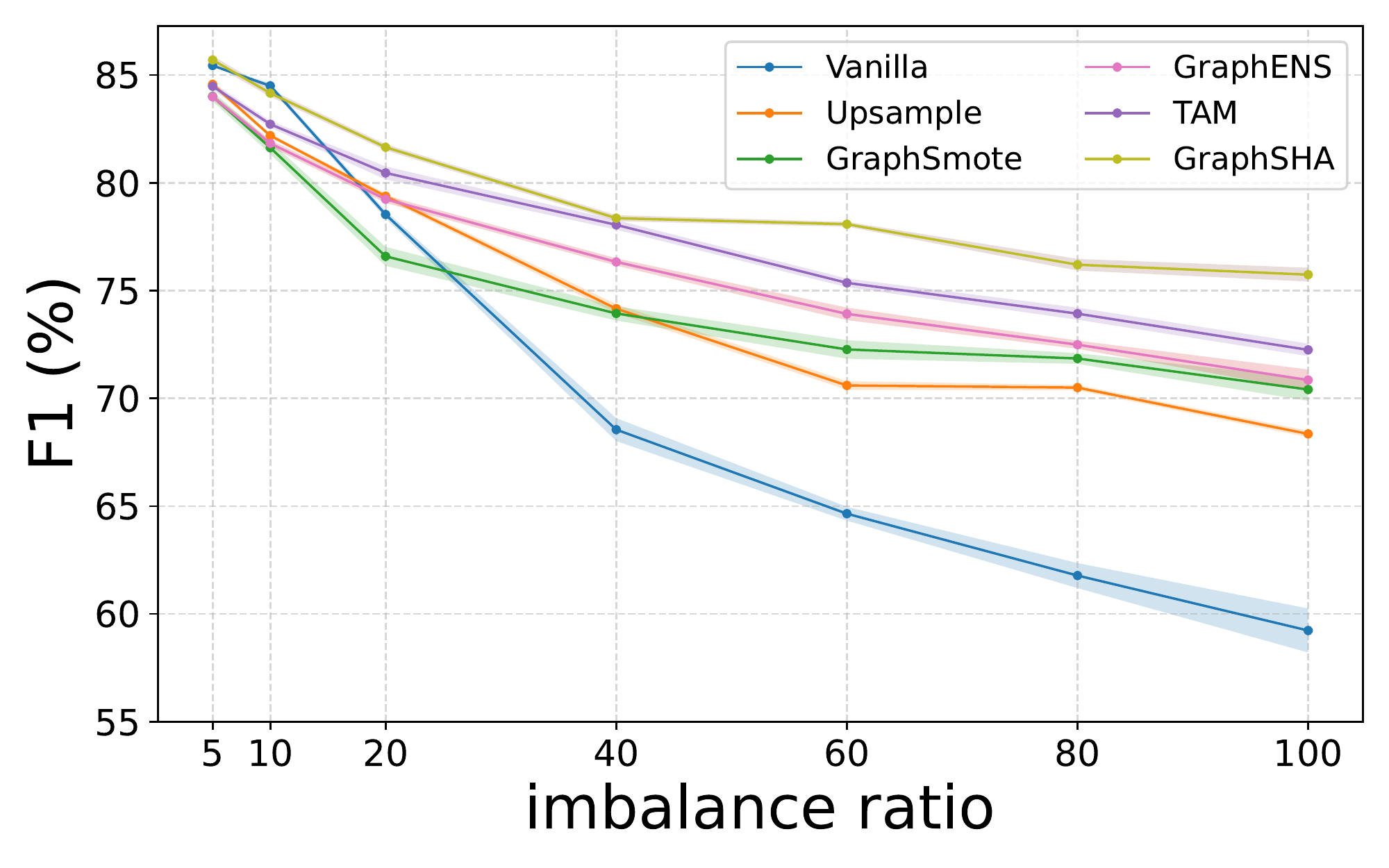}
\centering
\caption{Changing trend of F1-score with the increase of imbalance ratio on Cora-LT with GCN.}
\label{figure:trend}
\vspace{-0.4cm}
\end{figure}

\vspace{-0.2cm}
\subsection{Influence of Imbalance Ratio}
Furthermore, we analyze the performance of various baselines with different imbalance ratios $\rho$ from 5 to the extreme condition of 100, as shown in Figure~\ref{figure:trend} on Cora-LT with GCN backbone. We can see that the F1 scores of all methods are high when $\rho$ is small. With the increase of $\rho$, the performance of GraphSHA remains relatively stable, and its superiority increases when $\rho$ becomes larger, which indicates the effectiveness of GraphSHA in handling extreme class imbalance problem on graphs.

\subsection{Case Study}
We also present a case study on per-class accuracy for the baseline methods and GraphSHA with GCN backbone on Cora-LT in Table~\ref{table:per}. For generative approaches, GraphSmote only shows a tiny improvement for minor classes compared to Upsample, which verifies that synthesizing within minor classes could hardly enlarge the decision boundary. GraphENS, on the other hand, shows decent accuracy for minor classes. However, it is at the cost of the performance reduction for major classes, as the accuracy for $C_6$ is the lowest, which verifies that GraphENS overdoes the minor node generation. Our GraphSHA can avoid both problems as it shows superior accuracy for both minor and major classes, which benefits from the $\textsc{SemiMixup}$ module to synthesize harder minor samples to enlarge the minor decision boundaries effectively.

\begin{table}[!t]
\centering
\begin{center}
\caption{Classification accuracy for each class on Cora-LT.}\label{table:per}
\scalebox{0.8}{
\begin{tabular}{l|ccccccc}
\toprule
{\tabincell{c}{\textbf{Class} \\ \textbf{Distribution}}} & {\tabincell{c}{$C_0$ \\ (0.5\%)}} & {\tabincell{c}{$C_1$ \\ (1.1\%)}} & {\tabincell{c}{$C_2$ \\ (2.4\%)}} & {\tabincell{c}{$C_3$ \\(5.4\%)}} & {\tabincell{c}{$C_4$ \\ (11.6\%)}} & {\tabincell{c}{$C_5$ \\ (25.0\%)}} & {\tabincell{c}{$C_6$ \\ (54.0\%)}} \\
\midrule
Vanilla (GCN) & 0.0 & 28.6 & 67.0 & 60.0 & 81.2 & 93.8 & 93.1 \\
\midrule
Reweight & 32.9 & 70.9 & 75.0 & 67.5 & 78.6 & 93.5 & 87.9 \\
PC Softmax & 29.7 & 78.0 & 70.9 & 66.2 & 81.2 & 93.8 & 82.5 \\
CB Loss & 31.3 & 74.7 & 72.8 & 69.2 & 81.9 & 94.0 & 83.4 \\
Focal Loss & 29.9 & 75.9 & 72.2 & 71.9 & 82.6 & 94.6 & 83.2 \\
ReNode & 35.9 & 73.6 & 72.8 & 66.9 & 83.9 & 95.1 & 88.1 \\
\midrule
Upsample & 12.5 & 58.2 & 65.1 & 67.5 & 76.5 & 92.4 & 89.7 \\
GraphSmote & 22.2 & 66.4 & 68.9 & 62.1 & 79.4 & 93.4 & 89.4 \\
GraphENS & 37.7 & 72.2 & 73.2 & 63.5 & 74.6 & 94.7 & 82.3 \\
TAM (G-ENS) & 32.6 & 76.3 & 68.9 & 71.5 & 74.2 & 95.1 & 86.9 \\
\rowcolor{light-gray}\textbf{GraphSHA} & 51.6 & 76.9 & 66.0 & 65.4 & 76.5 & 93.8 & 92.1 \\
\bottomrule
\end{tabular}
}
\end{center}
\vspace{-0.3cm}
\end{table}

\subsection{Ablation Study}
In this subsection, we conduct a series of experiments to demonstrate the effectiveness of each component of GraphSHA, including how synthesizing harder minor samples and the \textsc{SemiMixup} module affect the model performance.
Specifically, we compare GraphSHA with several of its ablated variants starting from the vanilla GNN.
The results are shown in Table~\ref{table:ablation} on the Cora-LT dataset with GCN backbone. Here, easy samples are those that are far from the class boundary. Synthesizing easy samples is somewhat like GraphSmote as they both generate samples within class subspace, and we can see that they achieve similar results — only slightly better than the Upsample baseline.
Harder samples w/o \textsc{SemiMixup} are generated via the mixup between auxiliary nodes' 1-hop subgraph and the anchor nodes, i.e., substituting $\mathcal{N}_{syn}\sim P_{1hop}^{diff}(v_{anc})$ in Eq.~\eqref{eq:semihardsample} to $\mathcal{N}_{syn}\sim P_{1hop}^{diff}(v_{aux})$ to take a more adventurous step. We can see that the performance of minor classes improves. However, as we analyzed before, it is at the cost of degrading major classes — it performs the worst for the most major class.
On the other hand, Applying the \textsc{SemiMixup} module can alleviate the degradation while maintaining the accuracy for minor classes. Furthermore, GraphSHA can achieve better performance when weighted adjacency matrix is leveraged via Heat Kernel (HK) or Personal PageRank (PPR), which shows the importance of distinguishing the different importance of topological structure.

\subsection{Hyper-parameter Analysis}\label{sec:hyper}

\begin{figure}[!t]
\centering
\includegraphics[width=\linewidth]{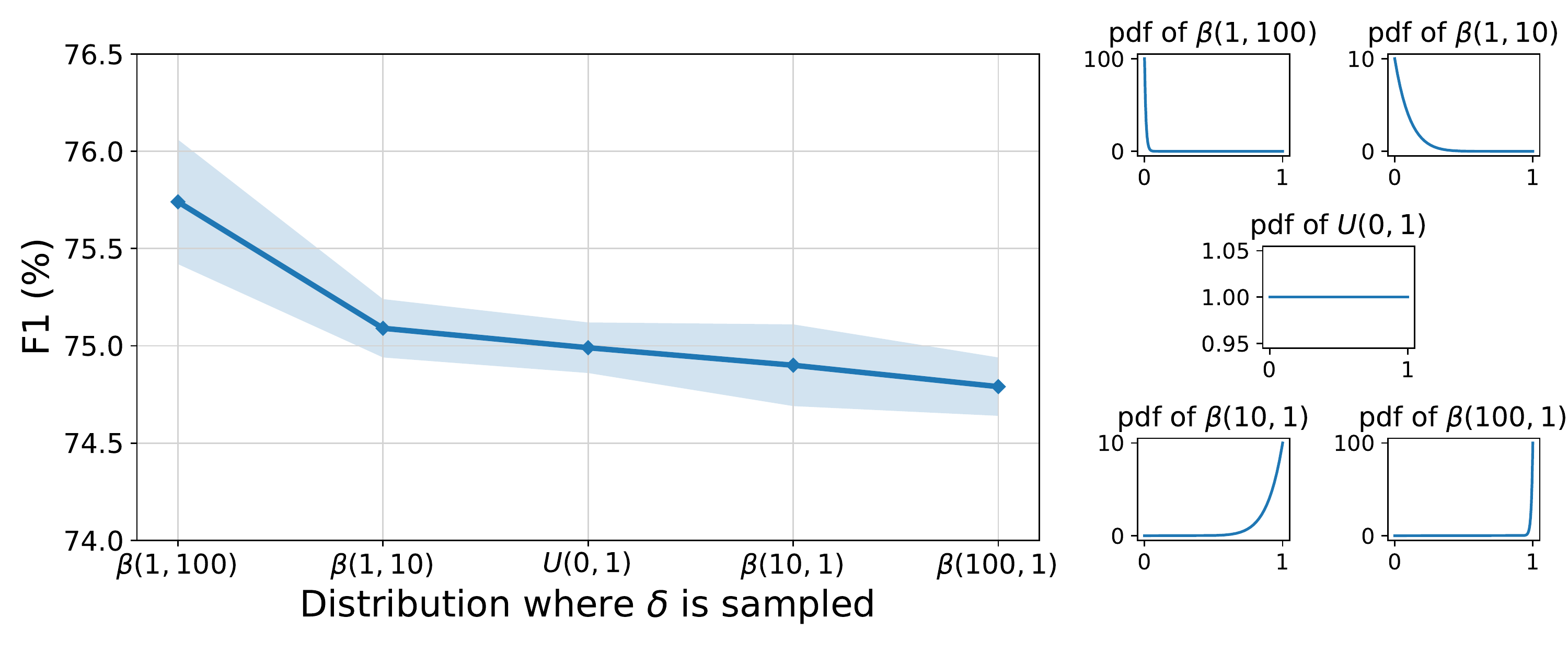}
\centering
\vspace{-0.3cm}
\caption{Left: Performance of GraphSHA w.r.t. different distributions where $\delta$ is sampled for $\boldsymbol{X}_{syn}=\delta\boldsymbol{X}_{anc}+(1-\delta)\boldsymbol{X}_{aux}$ on Cora-LT with GCN. Right: corresponding probability density functions (pdf) of the distributions.}
\label{figure:beta}
\end{figure}

In GraphSHA, random variable $\delta$ is used to control the hardness of the synthesized sample as $\boldsymbol{X}_{syn}=\delta\boldsymbol{X}_{anc}+(1-\delta)\boldsymbol{X}_{aux}$, and smaller $\delta$ indicates bias to harder samples. Here, we change the distribution where $\delta$ is sampled from, and the classification performance in terms of F1 is elaborated in Figure~\ref{figure:beta} on Cora-LT with GCN backbone. We can see that model performance drops as $\mathbb{E}(\delta)$ increases, which shows that synthesizing harder samples via the \textsc{SemiMixup} module is more beneficial for the model, as it can enlarge the minor subspaces to a greater extent.

\subsection{Analysis on Squeezed Minority Problem}

\begin{figure}[!t]
\centering
\includegraphics[width=\linewidth]{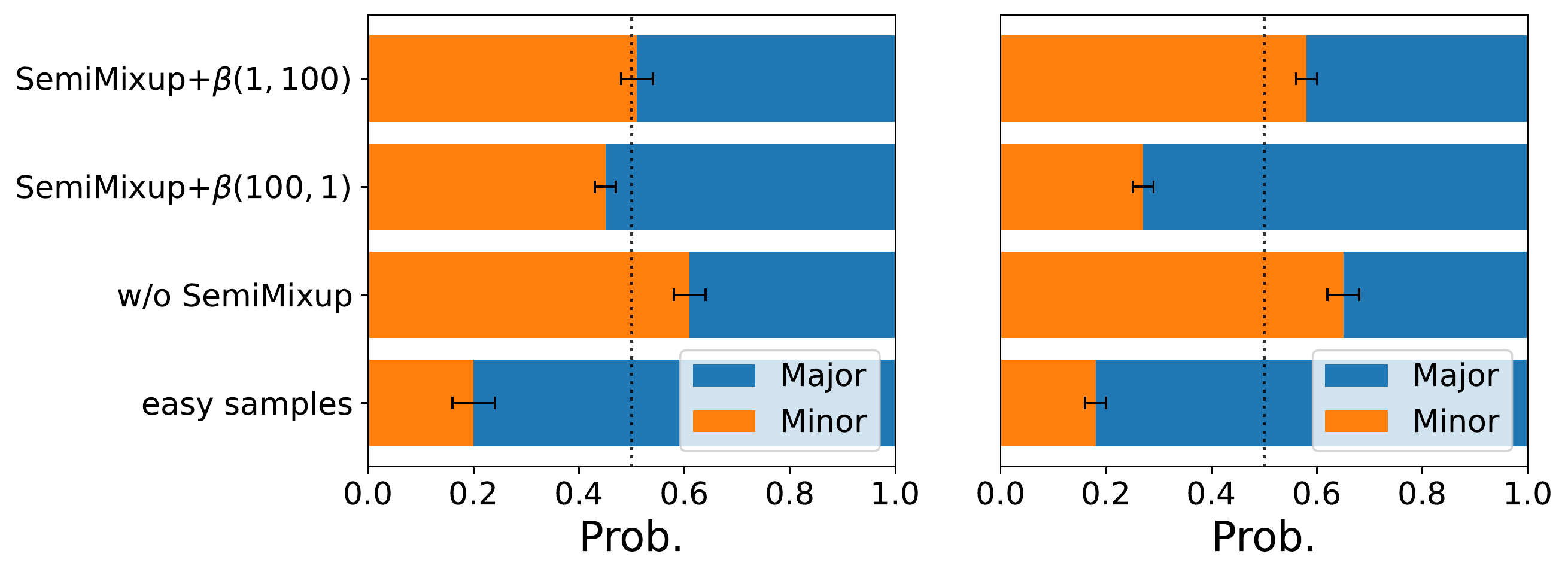}
\centering
\vspace{-0.2cm}
\caption{probability distribution of misclassified samples on Cora-LT (left) and CiteSeer-LT (right) with GCN backbone.}
\label{figure:response}
\vspace{-0.1cm}
\end{figure}

We conduct an experiment to validate how GraphSHA remits the squeezed minority problem by plotting the probability distribution of misclassified samples on Cora-LT and CiteSeer-LT, with the same setting as in Figure~\ref{figure:smp}. The result is shown in Figure~\ref{figure:response}, from which we have the following observations:

\begin{itemize}
\setlength{\leftskip}{-2em}
\item Synthesizing easy samples cannot tackle the squeezed minority problem, as these generated samples are confined in the raw latent spaces of minor classes.
\item Synthesizing harder samples without \textsc{SemiMixup} can tackle the squeezed minority problem by enlarging the minor subspaces to a large margin. However, it overdoes the enlargement as it is prone to classify major classes as minor ones.
\item Synthesizing via \textsc{SemiMixup} can remit the squeezed minority problem properly, as the probability of misclassified samples being minor classes is close to 0.5. Furthermore, the closer the synthesized node feature to the neighbor class (i.e., larger $\mathbb{E}(\delta)$), the better the problem is remitted, as the minor subspaces are enlarged to a greater extent.
This observation is consistent with the hyper-parameter analysis in Section~\ref{sec:hyper}.
\end{itemize}

\subsection{Visualization}

\begin{figure}[!t]
\centering
\subfigure[Training set samples]{
\centering
\includegraphics[width=0.22\textwidth]{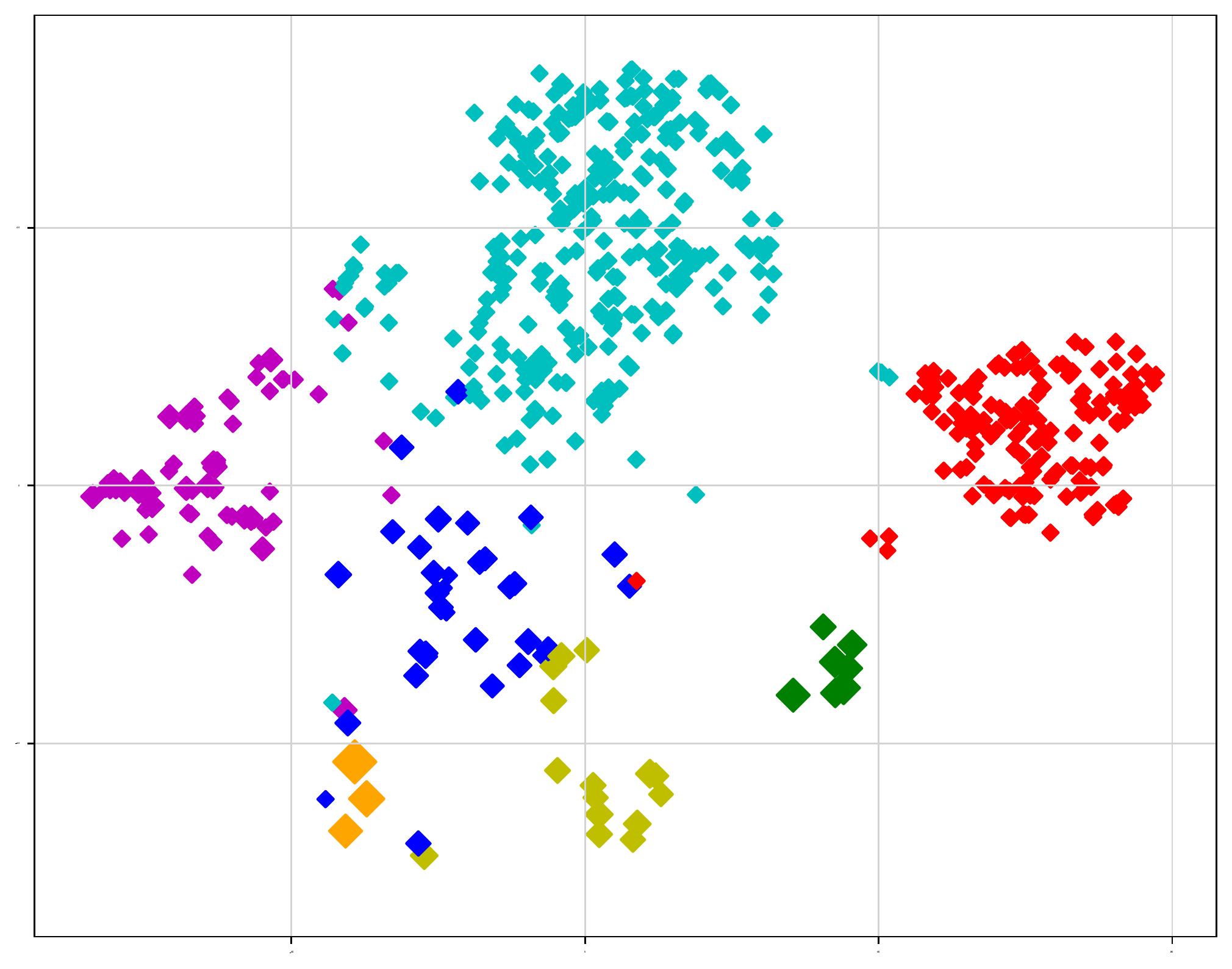}
}
\subfigure[Synthesized samples]{
\centering
\includegraphics[width=0.22\textwidth]{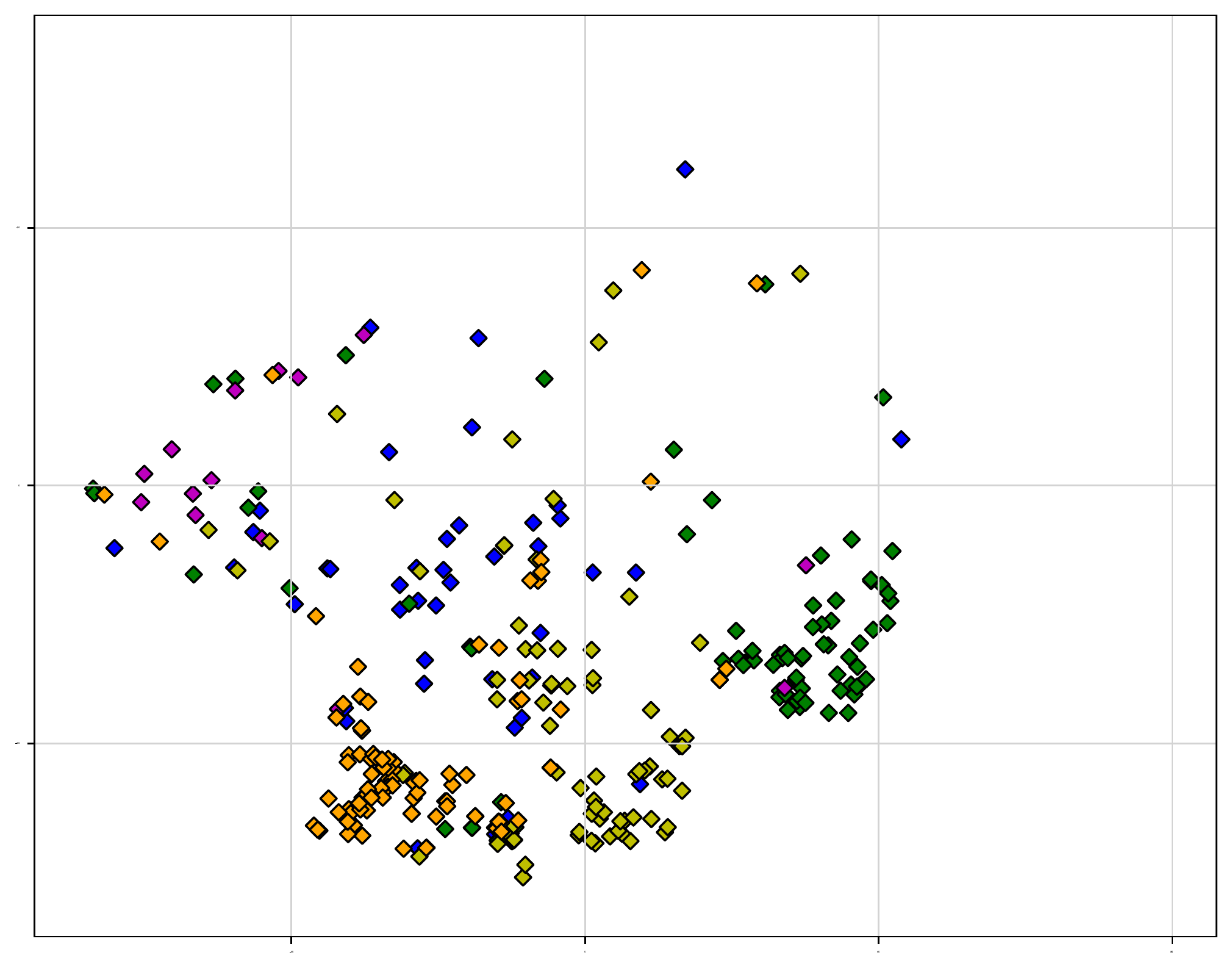}
}
\subfigure[Test set samples]{
\centering
\includegraphics[width=0.22\textwidth]{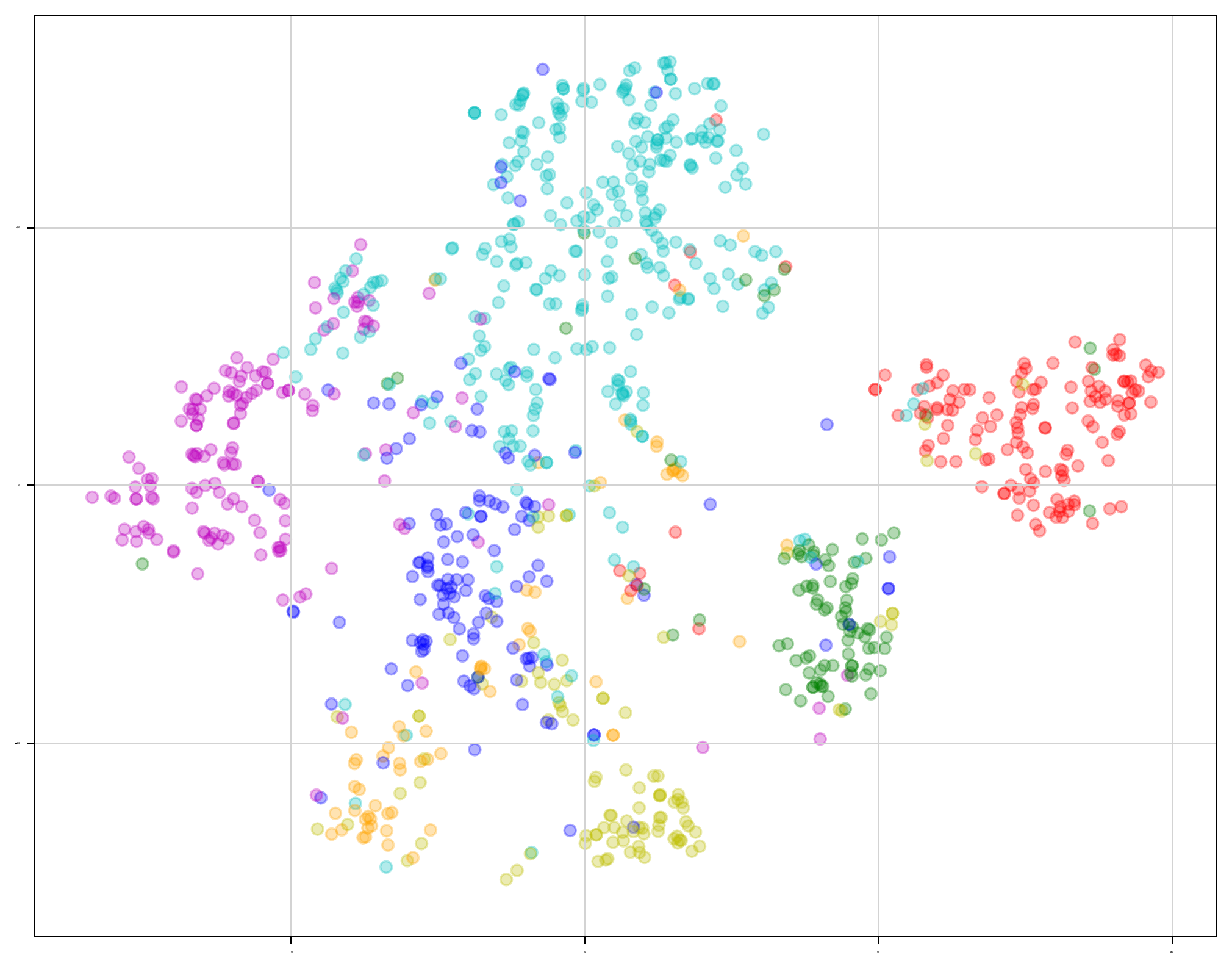}
}
\subfigure[All samples from (a), (b), and (c)]{
\centering
\includegraphics[width=0.22\textwidth]{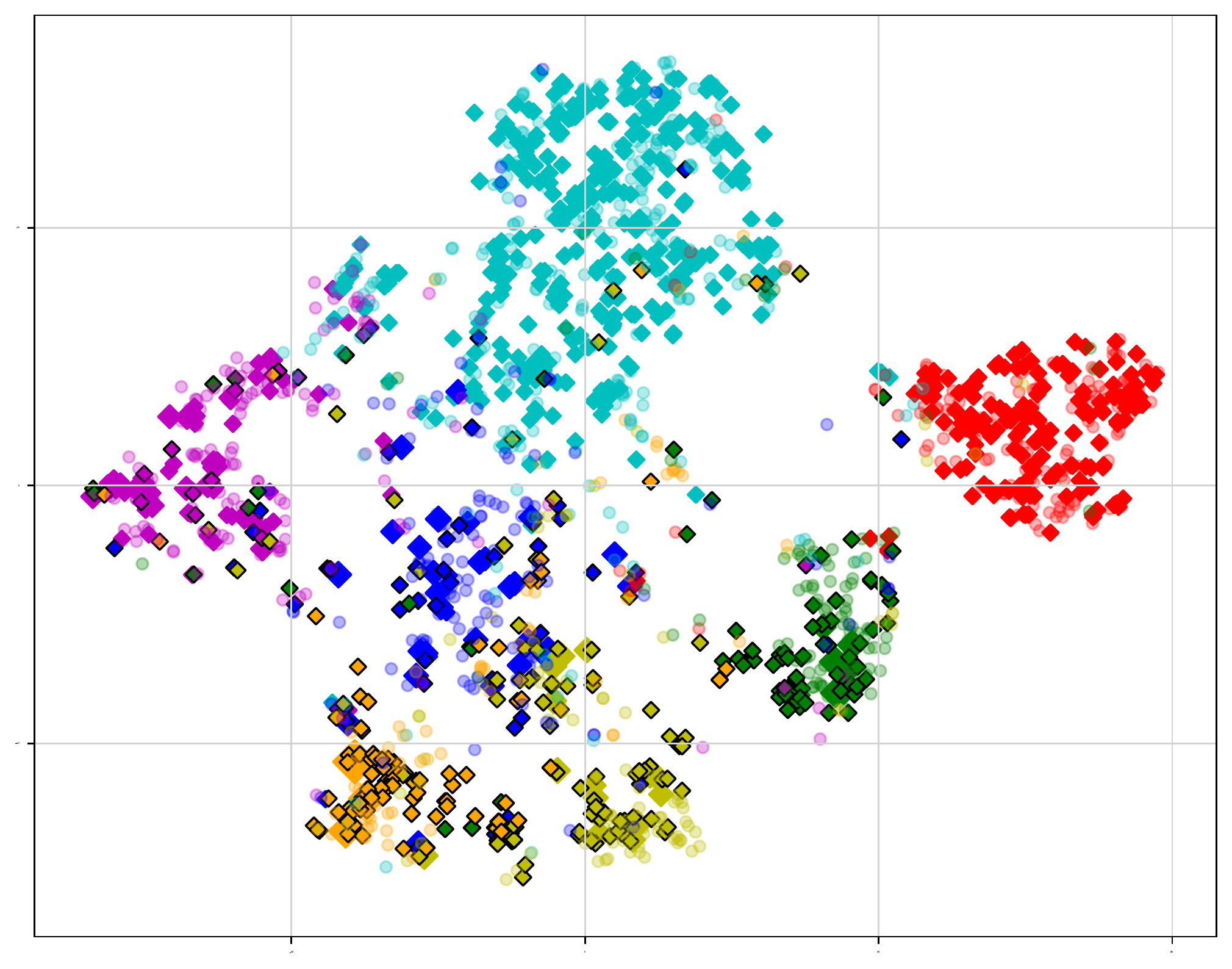}
}
\centering
\vspace{-0.2cm}
\caption{Visualization of GraphSHA on Cora-LT with GCN, where each node is colored by its label. In (a), the hardness of each training node is marked via the node size.}
\label{figure:visual}
\vspace{-0.3cm}
\end{figure}

We also provide an intuitive understanding of the synthesis of GraphSHA via t-SNE~\cite{tsne}-based visualization in Figure~\ref{figure:visual}, where each node is colored by its label in the latent space. From Figure~\ref{figure:visual}(a), we can see that the class-imbalanced graph suffers from the squeezed minority problem as the subspaces of minor classes are pretty small. And GraphSHA can recognize node hardness effectively as the nodes near the minor decision boundary are prone to be chosen as source nodes. In Figure~\ref{figure:visual}(b), GraphSHA can factually enlarge the minor class boundaries by synthesizing plausible harder minor nodes. As the enlarged minor subspaces could include minor samples in the test set to a large extent, GraphSHA can effectively remit the squeezed minority problem, as shown in Figure~\ref{figure:visual}(c), (d).

\section{Conclusion and Future Work}

In this paper, we study class-imbalanced node classification and find the squeezed minority problem, where the subspaces of minor classes are squeezed by major ones.
Inspired to enlarge the minor subspaces, we propose GraphSHA to synthesize harder minor samples with \textsc{semiMixup} module to avoid invading the subspaces of neighbor classes.
GraphSHA demonstrates superior performance on both manually and naturally imbalanced datasets compared against ten baselines with three GNN backbones. Furthermore, in-depth investigations also show the effectiveness of leveraging hard samples and the \textsc{semiMixup} module to enlarge minor class boundaries.
For future work, we expect GraphSHA to be generalized to other quantity-imbalanced scenarios on graphs like heterogeneous information network which has imbalanced node types.

\begin{acks}
The authors would like to thank Zhilin Zhao from University of Technology Sydney, Kunyu Lin from Sun Yat-sen University, and Dazhong Shen from University of Science and Technology of China for their insightful discussions.
This work was supported by NSFC (62276277), Guangdong Basic  Applied Basic Research Foundation (2022B1515120059), and the Foshan HKUST Projects (FSUST21-FYTRI01A, FSUST21-FYTRI02A).
Chang-Dong Wang and Hui Xiong are the corresponding authors.
\end{acks}

\bibliographystyle{ACM-Reference-Format}
\balance
\bibliography{sample-base}

\clearpage
\appendix

\section*{Appendix}

\section{Detailed Experimental Settings}
We introduce the detailed experimental settings, including the descriptions of datasets and implementation details for baselines in this section.

\subsection{Datasets}\label{appendix:datasets}
We adopt six widely-used datasets in the community, including Cora, CiteSeer, PubMed, Amazon-Photo, Amazon-Computers, and Coauthor-CS to conduct all the experiments throughout the paper. These datasets are collected from real-world scenarios of citation networks and co-purchase networks. Please note that all the datasets are accessible via PyTorch Geometric library~\cite{pyg}.

\begin{itemize}
\setlength{\leftskip}{-2em}
\item \textbf{Cora}, \textbf{CiteSeer}, and \textbf{PubMed}~\cite{cora} are three citation networks where nodes represent papers and edges represent citation relations. Each node in Cora and CiteSeer is described by a 0/1-valued word vector indicating the absence/presence of the corresponding word from the dictionary, while each node in PubMed is described by a TF/IDF weighted word vector from the dictionary. The nodes are categorized by their related research area for the three datasets. These datasets are accessible via \url{https://github.com/kimiyoung/planetoid/raw/master/data}.

\item \textbf{Amazon-Photo} and \textbf{Amazon-Computers}~\cite{amazon} are two co-purchase networks constructed from Amazon where nodes represent products and edges represent co-purchase relations.
Each node is described by a raw bag-of-words feature encoding product reviews and is labeled with its category. These datasets are accessible via \url{https://github.com/shchur/gnn-benchmark/raw/master/data/npz/}.

\item \textbf{Coauthor-CS}~\cite{amazon} is an academic network where nodes represent authors and edges represent co-author relations. Each node is described by a raw bag-of-words feature encoding keywords of his/her publication and is labeled with the most related research field. The dataset is accessible via \url{https://github.com/shchur/gnn-benchmark/raw/master/data/npz/}.

\item \textbf{ogbn-arXiv}~\cite{ogb} is a citation network between all Computer Science arXiv papers indexed by Microsoft academic graph~\cite{mag}, where nodes represent papers and edges represent citation relations. Each node is described by a 128-dimensional feature vector obtained by averaging the skip-gram word embeddings in its title and abstract. The nodes are categorized by their related research area.
According to the benchmark, the data split is based on the publication dates of the papers where the training set is papers published until 2017, the validation set is papers published in 2018, and the test set is papers published since 2019. The dataset is accessible via \url{https://ogb.stanford.edu/docs/nodeprop/#ogbn-arxiv}.
\end{itemize}

\subsection{Implementation Details for Baselines}\label{appendix:baselines}

We give detailed configurations for baseline methods. Please note we adopt same 2-layer GNN (GCN~\cite{gcn}, GAT~\cite{gat}, GraphSAGE~\cite{graphsage}) with hidden dimension 64 as the encoder for all the methods (except for arXiv where we set the hidden dimension as 256 to make a fair comparision on the leaderboard), including proposed GraphSHA, for a fair comparison.

For Reweight, the loss weight for each class is proportional to the number of samples.
For CB Loss, hyper-parameter $\beta$ is set to 0.999.
For Focal, hyper-parameter $\alpha$ is set to 2.0.
For ReNode, we combine the reweighting method for topological imbalance with Focal loss, and the lower and upper bounds for the topological imbalance are set to 0.5 and 1.5, respectively.
For Upsample, we duplicate minor nodes along with their edges until the number of each class sample reaches the mean number of class samples.
For GraphSmote, we choose the GraphSmote$_O$ variant, which predicts discrete values without pretraining, as it shows superior performance among multiple versions.
For GraphENS, we set the feature masking rate $k$ as 0.01 and temperature $\tau$ as 1, as suggested in the released codes.
For TAM, we choose the GraphENS-based version as it performs the best according to the paper, where the coefficient for ACM $\alpha$, the coefficient for ADM $\beta$, and the coefficient for classwise temperature $\phi$ are set to 2.5, 0.5, 1.2 respectively, which are the default settings in the released codes.

\section{Extra Experiments}\label{appendix:exp}
\vspace{0.2cm}

\begin{table}[!b]
\centering
\begin{center}
\caption{Node classification results ($\pm$std) on 
CoraFull which has 70 classes.}\label{table:corafull}
\vspace{-0.1cm}
\scalebox{1.00}{
\begin{tabular}{l|cc}
\toprule
$\rho$=100 & {Acc./bAcc.} & {F1} \\
\midrule
Vanilla (SAGE) & 56.40\footnotesize{$\pm$0.71} & 51.84\footnotesize{$\pm$0.78} \\
\cmidrule(lr){1-3}
Reweight & 68.17\footnotesize{$\pm$0.96} & 67.14\footnotesize{$\pm$0.96} \\
PC Softmax & 66.06\footnotesize{$\pm$0.38} & 64.74\footnotesize{$\pm$0.33} \\
CB Loss & \underline{69.89}\footnotesize{$\pm$1.11} & \underline{69.18}\footnotesize{$\pm$1.21} \\
Focal Loss & 67.83\footnotesize{$\pm$1.20} & 66.54\footnotesize{$\pm$1.33} \\
ReNode & 59.52\footnotesize{$\pm$0.58} & 57.94\footnotesize{$\pm$0.52} \\
\cmidrule(lr){1-3}
Upsample & 63.14\footnotesize{$\pm$0.77} & 61.70\footnotesize{$\pm$0.72} \\
GraphSmote & 58.86\footnotesize{$\pm$0.59} & 56.95\footnotesize{$\pm$0.66} \\
GraphENS & 63.14\footnotesize{$\pm$0.92} & 62.28\footnotesize{$\pm$0.88} \\
TAM (G-ENS) & 64.86\footnotesize{$\pm$0.78} & 63.73\footnotesize{$\pm$0.72} \\
\rowcolor{light-gray}\textbf{GraphSHA} & \textbf{73.43}\footnotesize{$\pm$0.42} & \textbf{72.50}\footnotesize{$\pm$0.43} \\
\bottomrule
\end{tabular}
}
\end{center}
\end{table}

We first present an extra experiment on CoraFull~\cite{corafullnew} which has a larger class set (70 classes). As the dataset follows a long-tailed distribution intrinsically, we sample the same number of nodes from each class for validation/test sets randomly and assign the remaining nodes as the training set. The imbalance ratio is set to 100 and GraphSAGE is adopted as the backbone encoder. 
As the test set is balanced, the Acc. and bAcc. metrics are the same.
The result is shown in Table~\ref{table:corafull}, from which We can see that the superiority of GraphSHA is still significant, which shows the generalization and effectiveness of GraphSHA in handling class-imbalanced node classification tasks.

\begin{table*}[!t]
\centering
\begin{center}
\caption{Node classification results ($\pm$std) on Photo, Computer, and CS in step class-imbalanced setting with GCN and GAT backbones.}\label{table:appstep}
\scalebox{0.99}{
\begin{tabular}{cl|ccc|ccc|ccc}
\toprule
\multirow{2}{*}{} & \textbf{Dataset} & \multicolumn{3}{c|}{Photo-ST} & \multicolumn{3}{c|}{Computer-ST} & \multicolumn{3}{c}{CS-ST} \\
\cmidrule(lr){2-2}\cmidrule(lr){3-5}\cmidrule(lr){6-8}\cmidrule(lr){9-11}
& $\rho$=20 & {Acc.} & {bAcc.} & {F1} & {Acc.} & {bAcc.} & {F1} & {Acc.} & {bAcc.} & {F1} \\
\midrule
\midrule
\multirow{11}{*}{\rotatebox{90}{GCN}}
& Vanilla & 37.79\footnotesize{$\pm$0.22} & 46.77\footnotesize{$\pm$0.11} & 27.15\footnotesize{$\pm$0.43} & 56.12\footnotesize{$\pm$1.41} & 41.49\footnotesize{$\pm$0.63} & 27.76\footnotesize{$\pm$0.53} & 37.36\footnotesize{$\pm$0.97} & 54.35\footnotesize{$\pm$0.72} & 30.47\footnotesize{$\pm$1.19} \\
\cmidrule(lr){2-11}
& Reweight & 85.81\footnotesize{$\pm$0.13} & 88.62\footnotesize{$\pm$0.06} & 83.30\footnotesize{$\pm$0.14} & 78.77\footnotesize{$\pm$0.25} & 85.30\footnotesize{$\pm$0.10} & 74.31\footnotesize{$\pm$0.23} & 91.86\footnotesize{$\pm$0.06} & 91.62\footnotesize{$\pm$0.06} & 82.46\footnotesize{$\pm$0.15} \\
& PC Softmax & 64.66\footnotesize{$\pm$1.73} & 71.56\footnotesize{$\pm$1.16} & 61.31\footnotesize{$\pm$1.25} & 73.33\footnotesize{$\pm$1.22} & 60.07\footnotesize{$\pm$1.82} & 55.09\footnotesize{$\pm$2.27} & 87.38\footnotesize{$\pm$0.49} & 87.46\footnotesize{$\pm$0.39} & 74.24\footnotesize{$\pm$0.77} \\
& CB Loss & 86.85\footnotesize{$\pm$0.05} & 88.69\footnotesize{$\pm$0.05} & \underline{84.78}\footnotesize{$\pm$0.12} & \textbf{82.22}\footnotesize{$\pm$0.13} & 86.71\footnotesize{$\pm$0.05} & \underline{75.80}\footnotesize{$\pm$0.13} & 91.43\footnotesize{$\pm$0.05} & 91.25\footnotesize{$\pm$0.07} & 77.72\footnotesize{$\pm$0.85} \\
& Focal Loss & 86.14\footnotesize{$\pm$0.17} & 88.44\footnotesize{$\pm$0.11} & 84.12\footnotesize{$\pm$0.23} & 81.01\footnotesize{$\pm$0.19} & \textbf{86.89}\footnotesize{$\pm$0.07} & 75.50\footnotesize{$\pm$0.17} & 91.01\footnotesize{$\pm$0.08} & 90.72\footnotesize{$\pm$0.04} & 79.80\footnotesize{$\pm$0.77} \\
& ReNode & 86.08\footnotesize{$\pm$0.18} & 87.34\footnotesize{$\pm$0.34} & 82.51\footnotesize{$\pm$0.29} & 72.92\footnotesize{$\pm$0.97} & 78.12\footnotesize{$\pm$0.84} & 67.04\footnotesize{$\pm$1.13} & 92.02\footnotesize{$\pm$0.21} & 91.08\footnotesize{$\pm$0.19} & 82.87\footnotesize{$\pm$0.97} \\
\cmidrule(lr){2-11}
& Upsample & 85.40\footnotesize{$\pm$0.18} & 87.32\footnotesize{$\pm$0.15} & 82.79\footnotesize{$\pm$0.22} & 80.07\footnotesize{$\pm$0.31} & 85.10\footnotesize{$\pm$0.11} & 74.85\footnotesize{$\pm$0.21} & 86.11\footnotesize{$\pm$0.14} & 86.82\footnotesize{$\pm$0.10} & 75.55\footnotesize{$\pm$0.13} \\
& GraphSmote & 83.99\footnotesize{$\pm$0.20} & 86.53\footnotesize{$\pm$0.19} & 81.86\footnotesize{$\pm$0.21} & 76.76\footnotesize{$\pm$0.18} & 84.10\footnotesize{$\pm$0.17} & 69.40\footnotesize{$\pm$0.19} & 86.20\footnotesize{$\pm$0.17} & 85.44\footnotesize{$\pm$0.15} & 69.04\footnotesize{$\pm$0.64} \\
& GraphENS & \underline{87.00}\footnotesize{$\pm$0.07} & \textbf{89.19}\footnotesize{$\pm$0.06} & 84.66\footnotesize{$\pm$0.09} & 79.71\footnotesize{$\pm$0.08} & 86.50\footnotesize{$\pm$0.08} & 74.55\footnotesize{$\pm$0.10} & \underline{92.17}\footnotesize{$\pm$0.10} & \underline{91.94}\footnotesize{$\pm$0.11} & 82.90\footnotesize{$\pm$0.43} \\
& TAM (G-ENS) & 84.37\footnotesize{$\pm$0.11} & 86.41\footnotesize{$\pm$0.09} & 81.91\footnotesize{$\pm$0.10} & 76.26\footnotesize{$\pm$0.23} & 83.38\footnotesize{$\pm$0.26} & 73.85\footnotesize{$\pm$0.22} & 92.15\footnotesize{$\pm$0.22} & 91.92\footnotesize{$\pm$0.24} & \underline{83.13}\footnotesize{$\pm$0.53} \\
& \cellcolor{light-gray}\textbf{GraphSHA} & \cellcolor{light-gray}\textbf{87.40}\footnotesize{$\pm$0.09} & \cellcolor{light-gray}\underline{88.92}\footnotesize{$\pm$0.09} & \cellcolor{light-gray}\textbf{85.18}\footnotesize{$\pm$0.11} & \cellcolor{light-gray}\underline{81.75}\footnotesize{$\pm$0.14} & \cellcolor{light-gray}\underline{86.75}\footnotesize{$\pm$0.09} & \cellcolor{light-gray}\textbf{76.86}\footnotesize{$\pm$0.30} & \cellcolor{light-gray}\textbf{92.38}\footnotesize{$\pm$0.09} & \cellcolor{light-gray}\textbf{92.01}\footnotesize{$\pm$0.06} & \cellcolor{light-gray}\textbf{83.33}\footnotesize{$\pm$0.45} \\
\midrule
\midrule
\multirow{11}{*}{\rotatebox{90}{GAT}}
& Vanilla & 37.54\footnotesize{$\pm$0.34} & 45.95\footnotesize{$\pm$0.32} & 28.87\footnotesize{$\pm$0.49} & 58.00\footnotesize{$\pm$0.69} & 42.82\footnotesize{$\pm$0.39} & 26.79\footnotesize{$\pm$0.19} & 34.48\footnotesize{$\pm$0.42} & 50.08\footnotesize{$\pm$0.65} & 24.92\footnotesize{$\pm$1.00} \\
\cmidrule(lr){2-11}
& Reweight & 80.34\footnotesize{$\pm$1.02} & 83.08\footnotesize{$\pm$0.50} & 76.64\footnotesize{$\pm$0.92} & 72.65\footnotesize{$\pm$0.40} & 76.81\footnotesize{$\pm$0.37} & 64.00\footnotesize{$\pm$0.35} & 88.31\footnotesize{$\pm$0.38} & 87.33\footnotesize{$\pm$0.39} & 71.67\footnotesize{$\pm$0.59} \\
& PC Softmax & 51.74\footnotesize{$\pm$3.22} & 61.48\footnotesize{$\pm$1.90} & 51.17\footnotesize{$\pm$2.44} & 31.56\footnotesize{$\pm$2.89} & 51.83\footnotesize{$\pm$2.52} & 37.70\footnotesize{$\pm$2.25} & 78.84\footnotesize{$\pm$0.58} & 77.80\footnotesize{$\pm$0.61} & 65.46\footnotesize{$\pm$0.64} \\
& CB Loss & 82.82\footnotesize{$\pm$0.79} & 86.44\footnotesize{$\pm$0.58} & 79.57\footnotesize{$\pm$0.93} & 79.60\footnotesize{$\pm$0.71} & 84.78\footnotesize{$\pm$0.23} & 74.11\footnotesize{$\pm$0.66} & 89.74\footnotesize{$\pm$0.26} & 89.68\footnotesize{$\pm$0.22} & 75.00\footnotesize{$\pm$0.83} \\
& Focal Loss & 83.03\footnotesize{$\pm$0.58} & 85.86\footnotesize{$\pm$0.43} & 79.39\footnotesize{$\pm$0.73} & 79.49\footnotesize{$\pm$0.45} & 85.04\footnotesize{$\pm$0.23} & 74.10\footnotesize{$\pm$0.48} & 88.73\footnotesize{$\pm$0.20} & 88.03\footnotesize{$\pm$0.22} & 73.08\footnotesize{$\pm$0.73} \\
& ReNode & 76.49\footnotesize{$\pm$1.00} & 81.35\footnotesize{$\pm$0.95} & 73.33\footnotesize{$\pm$0.86} & 71.71\footnotesize{$\pm$0.65} & 75.20\footnotesize{$\pm$0.36} & 63.94\footnotesize{$\pm$0.77} & 87.86\footnotesize{$\pm$0.29} & 85.55\footnotesize{$\pm$0.34} & 69.80\footnotesize{$\pm$0.47} \\
\cmidrule(lr){2-11}
& Upsample & 77.89\footnotesize{$\pm$0.83} & 81.16\footnotesize{$\pm$0.33} & 73.91\footnotesize{$\pm$0.59} & 74.86\footnotesize{$\pm$0.69} & 78.18\footnotesize{$\pm$0.45} & 66.28\footnotesize{$\pm$0.81} & 82.23\footnotesize{$\pm$0.18} & 82.70\footnotesize{$\pm$0.10} & 65.74\footnotesize{$\pm$0.17} \\
& GraphSmote & 80.71\footnotesize{$\pm$0.33} & 81.48\footnotesize{$\pm$0.38} & 76.96\footnotesize{$\pm$0.30} & 79.38\footnotesize{$\pm$0.25} & 84.66\footnotesize{$\pm$0.20} & 70.75\footnotesize{$\pm$0.27} & 83.46\footnotesize{$\pm$0.18} & 82.75\footnotesize{$\pm$0.18} & 67.02\footnotesize{$\pm$0.22} \\
& GraphENS & \textbf{84.22}\footnotesize{$\pm$0.36} & \underline{86.45}\footnotesize{$\pm$0.19} & \underline{80.02}\footnotesize{$\pm$0.30} & \textbf{80.78}\footnotesize{$\pm$0.18} & \underline{84.82}\footnotesize{$\pm$0.19} & \textbf{75.13}\footnotesize{$\pm$0.43} & 89.93\footnotesize{$\pm$0.30} & 89.71\footnotesize{$\pm$0.29} & \underline{79.66}\footnotesize{$\pm$0.38} \\
& TAM (G-ENS) & 80.94\footnotesize{$\pm$0.42} & 83.09\footnotesize{$\pm$0.36} & 78.89\footnotesize{$\pm$0.45} & 77.68\footnotesize{$\pm$0.24} & 82.97\footnotesize{$\pm$0.18} & \underline{74.22}\footnotesize{$\pm$0.39} & \textbf{91.86}\footnotesize{$\pm$0.36} &  \underline{90.96}\footnotesize{$\pm$0.33} & \textbf{80.41}\footnotesize{$\pm$0.35} \\
& \cellcolor{light-gray}\textbf{GraphSHA} & \cellcolor{light-gray}\underline{84.09}\footnotesize{$\pm$0.90} & \cellcolor{light-gray}\textbf{86.61}\footnotesize{$\pm$0.81} & \cellcolor{light-gray}\textbf{80.85}\footnotesize{$\pm$0.72} & \cellcolor{light-gray}\underline{80.01}\footnotesize{$\pm$0.42} & \cellcolor{light-gray}\textbf{84.89}\footnotesize{$\pm$0.27} & \cellcolor{light-gray}71.64\footnotesize{$\pm$0.55} & \cellcolor{light-gray}\underline{91.79}\footnotesize{$\pm$0.13} & \cellcolor{light-gray}\textbf{91.46}\footnotesize{$\pm$0.10} & \cellcolor{light-gray}76.66\footnotesize{$\pm$0.18} \\
\bottomrule
\end{tabular}
}
\end{center}
\end{table*}

\begin{table*}[!t]
\centering
\begin{center}
\caption{Node classification results ($\pm$std) on Cora, CiteSeer, and PubMed in long-tailed class-imbalanced setting with $K$NN-based node hardness.}\label{table:knn1}
\scalebox{0.99}{
\begin{tabular}{l|ccc|ccc|ccc}
\toprule
\textbf{Dataset} & \multicolumn{3}{c|}{Cora-LT} & \multicolumn{3}{c|}{CiteSeer-LT} & \multicolumn{3}{c}{PubMed-LT} \\
\cmidrule(lr){1-1}\cmidrule(lr){2-4}\cmidrule(lr){5-7}\cmidrule(lr){8-10}
$\rho$=100 & {Acc.} & {bAcc.} & {F1} & {Acc.} & {bAcc.} & {F1} & {Acc.} & {bAcc.} & {F1} \\
\midrule
\textbf{GraphSHA+GCN} & 79.21\footnotesize{$\pm$0.12} & 74.13\footnotesize{$\pm$0.15} & 75.21\footnotesize{$\pm$0.13} & 64.00\footnotesize{$\pm$0.29} & 58.82\footnotesize{$\pm$0.28} & 58.80\footnotesize{$\pm$0.31} & 78.00\footnotesize{$\pm$0.16} & 72.98\footnotesize{$\pm$0.27} & 73.43\footnotesize{$\pm$0.25} \\
\textbf{GraphSHA+GAT} & 78.87\footnotesize{$\pm$0.25} & 74.06\footnotesize{$\pm$0.34} & 74.89\footnotesize{$\pm$0.27} & 64.03\footnotesize{$\pm$0.32} & 58.29\footnotesize{$\pm$0.31} & 57.49\footnotesize{$\pm$0.30} & 78.86\footnotesize{$\pm$0.25} & 73.97\footnotesize{$\pm$0.25} & 74.92\footnotesize{$\pm$0.27} \\
\textbf{GraphSHA+SAGE} & 78.72\footnotesize{$\pm$0.25} & 73.36\footnotesize{$\pm$0.35} & 74.20\footnotesize{$\pm$0.22} & 63.98\footnotesize{$\pm$0.28} & 58.62\footnotesize{$\pm$0.28} & 58.27\footnotesize{$\pm$0.31} & 78.66\footnotesize{$\pm$0.29} & 73.71\footnotesize{$\pm$0.33} & 74.89\footnotesize{$\pm$0.30} \\
\bottomrule
\end{tabular}
}
\end{center}
\end{table*}

\begin{table*}[!t]
\centering
\begin{center}
\caption{Node classification results ($\pm$std) on Photo, Computer, and CS in step class-imbalanced setting with $K$NN-based node hardness.}\label{table:knn2}
\scalebox{0.99}{
\begin{tabular}{l|ccc|ccc|ccc}
\toprule
\textbf{Dataset} & \multicolumn{3}{c|}{Photo-ST} & \multicolumn{3}{c|}{Computer-ST} & \multicolumn{3}{c}{CS-ST} \\
\cmidrule(lr){1-1}\cmidrule(lr){2-4}\cmidrule(lr){5-7}\cmidrule(lr){8-10}
$\rho$=20 & {Acc.} & {bAcc.} & {F1} & {Acc.} & {bAcc.} & {F1} & {Acc.} & {bAcc.} & {F1} \\
\midrule
\textbf{GraphSHA+GCN} & 87.32\footnotesize{$\pm$0.13} & 88.96\footnotesize{$\pm$0.07} & 85.09\footnotesize{$\pm$0.15} & 81.56\footnotesize{$\pm$0.26} & 86.79\footnotesize{$\pm$0.10} & 76.96\footnotesize{$\pm$0.25} & 92.19\footnotesize{$\pm$0.18} & 92.08\footnotesize{$\pm$0.15} & 83.12\footnotesize{$\pm$0.33} \\
\textbf{GraphSHA+GAT} & 84.32\footnotesize{$\pm$0.28} & 86.62\footnotesize{$\pm$0.20} & 80.78\footnotesize{$\pm$0.30} & 80.05\footnotesize{$\pm$0.43} & 84.85\footnotesize{$\pm$0.33} & 71.63\footnotesize{$\pm$0.52} & 91.60\footnotesize{$\pm$0.25} & 91.26\footnotesize{$\pm$0.27} & 76.18\footnotesize{$\pm$0.32} \\
\textbf{GraphSHA+SAGE} & 89.04\footnotesize{$\pm$0.18} & 90.64\footnotesize{$\pm$0.18} & 87.27\footnotesize{$\pm$0.20} & 84.61\footnotesize{$\pm$0.18} & 89.40\footnotesize{$\pm$0.08} & 78.13\footnotesize{$\pm$0.27} & 93.05\footnotesize{$\pm$0.08} & 92.76\footnotesize{$\pm$0.12} & 82.47\footnotesize{$\pm$0.24} \\
\bottomrule
\end{tabular}
}
\end{center}
\end{table*}

We also present experiments in the step class imbalance setting with GCN and GAT as GNN backbones. The result is shown in Table~\ref{table:appstep}, from which we can see that GraphSHA still achieves the best performance overall.

In addation to confidence-based node hardness discussed in Section~\ref{sec:identifying}, we also consider $K$NN-based hardness, which is defined as $\mathcal{H}_i=\lvert\{v_j\vert \boldsymbol{H}_j\in\mathrm{RF}_{k}(\boldsymbol{H}_i),\boldsymbol{Y}(i)\neq\boldsymbol{Y}(j)\}\rvert/k$ where $\mathrm{RF}_{k}(\boldsymbol{H}_i)$ is the $K$NN receptive field for node $v_i$ in the latent space. For sampling $v_{anc}$ and $v_{aux}$ near the minor decision boundary, we can get $v_{anc}$ according to node hardness as in Section~\ref{sec:identifying}, and sample $v_{aux}$ in $\mathrm{RF}_{k}(\boldsymbol{H}_{anc})$ directly. The experimental results are shown in Table~\ref{table:knn1} for the LT setting and Table~\ref{table:knn2} for the ST setting, respectively. We can see that leveraging $K$NN-based hardness can also achieve remarkable performance, which validates that GraphSHA is agnostic to the way of calculating node hardness.

\end{document}